\pdfoutput=1

\documentclass[11pt]{article}

\usepackage[]{acl}

\usepackage{times}
\usepackage{latexsym}
\usepackage{hyperref}
\usepackage{booktabs}

\usepackage{enumitem}
\usepackage{longtable}
\usepackage{makecell} 
\usepackage{multirow}
\usepackage{float}

\usepackage[T1]{fontenc}

\usepackage[utf8]{inputenc}

\usepackage{microtype}

\usepackage{inconsolata}

\usepackage{graphicx}
\usepackage{fdsymbol}
\usepackage{rotating}

\makeatletter
\newcommand\blfootnote[1]{%
  \begingroup
  \renewcommand{\@makefntext}[1]{\noindent\makebox[1.8em][r]#1}
  \renewcommand\thefootnote{}\footnote{#1}%
  \addtocounter{footnote}{-1}%
  \endgroup
}
\makeatother

\title{Tell Me What You Know About Sexism: \\
Expert-LLM Interaction Strategies and Co-Created Definitions \\ for Zero-Shot Sexism Detection}

\author{
 \textbf{Myrthe Reuver{$\varheartsuit$}\textsuperscript{1}},
 \textbf{Indira Sen\textsuperscript{2}},
 \textbf{Matteo Melis\textsuperscript{3}},
 \textbf{Gabriella Lapesa\textsuperscript{3,4}}
\\
 \textsuperscript{1}Populytics
 \textsuperscript{2}University of Mannheim\\
 \textsuperscript{3}GESIS - Leibniz Institute for the Social Sciences  
\textsuperscript{4}Heinrich-Heine University Düsseldorf\\
\textsuperscript{1}\texttt{myrthe@populytics.nl}, \textsuperscript{2}\texttt{indira.sen@uni-mannheim.de},
\textsuperscript{3}\texttt{first.last@gesis.org} 
}
 
\begin{document}
\maketitle
\begin{abstract}

This paper investigates \textit{hybrid intelligence} and collaboration between researchers of sexism and Large Language Models (LLMs), with a four-component pipeline.
First, nine sexism researchers answer questions about their knowledge of sexism and of LLMs. They then participate in two interactive experiments involving an LLM (\texttt{GPT3.5}). The first experiment has experts assessing the model's knowledge about sexism and suitability for use in research. The second experiment tasks them with creating three different definitions of sexism: an expert-written definition, an LLM-written one, and a co-created definition. Lastly, zero-shot classification experiments use the three definitions from each expert in a prompt template for sexism detection, evaluating \texttt{GPT4o} on 2.500 texts sampled from five sexism benchmarks. We then analyze the resulting 67.500 classification decisions. The LLM interactions lead to longer and more complex definitions of sexism. Expert-written definitions on average perform poorly compared to LLM-generated definitions. However, some experts do improve classification performance with their co-created definitions of sexism, also experts who are inexperienced in using LLMs.

\end{abstract}

\blfootnote{
<$\varheartsuit$>
Majority of work done while employed at the Vrije Universiteit Amsterdam and visiting the Computational Social Science department at GESIS on a GESIS Visiting Junior Researcher grant.
}

\section{Introduction}\label{sec:intro}

    \begin{figure}[t]
    \centering \includegraphics[scale=0.18]{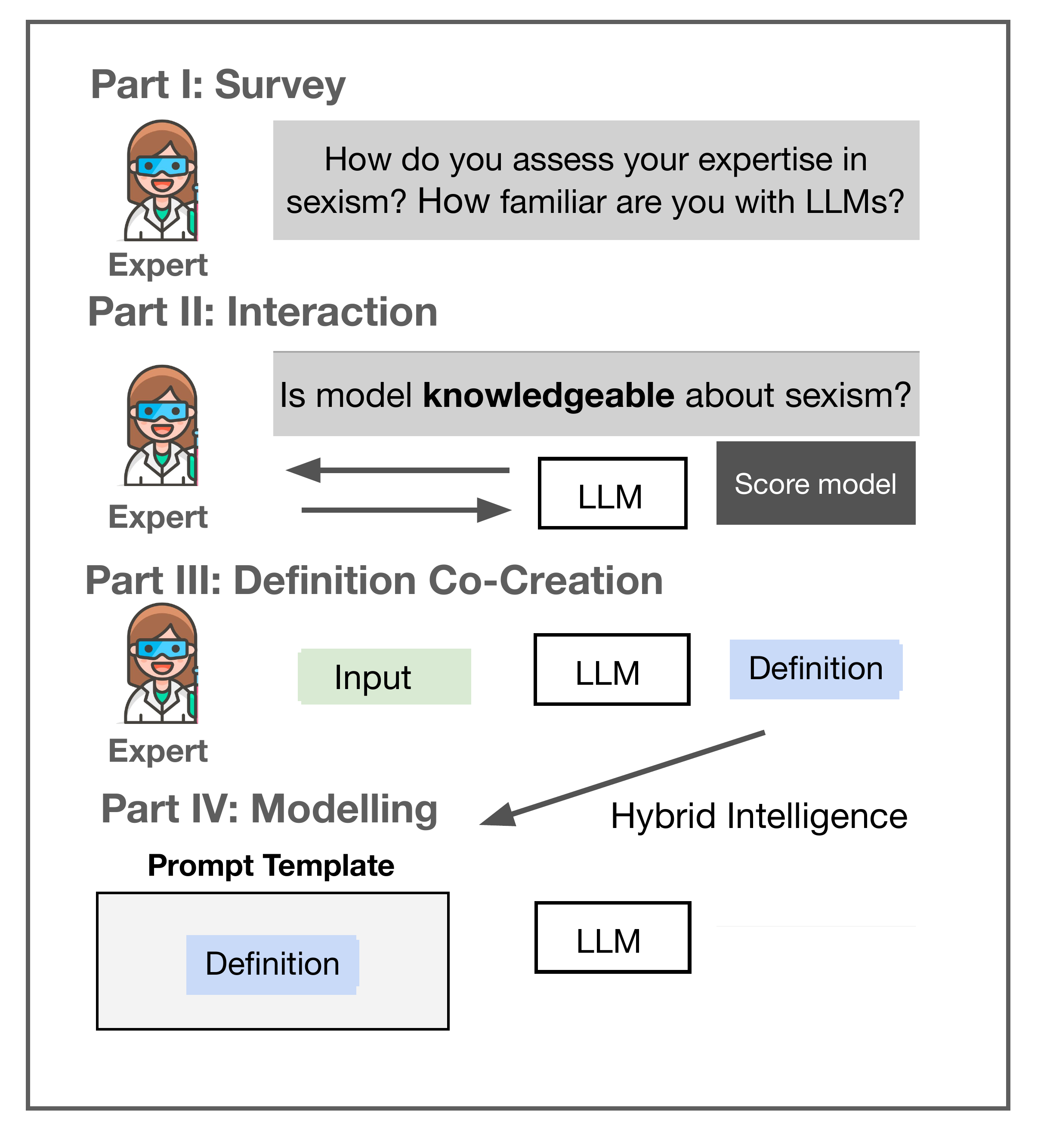}
    \caption{
  Experts participate in a survey (part I) as well as two interactive experiments (part II and III), after which we perform zero-shot classification experiments (part IV) with two LLMs, using the sexism definitions created during the interaction experiments.}
    \label{fig:process}
\end{figure}

Large Language Models (LLMs) with chat interfaces are an increasingly popular tool in
various scientific fields, for a variety of tasks: from writing assistance to data
annotation and data analysis. These interactive models produce pleasant and convincing (while not necessarily factually correct) conversations \cite{ji2023survey}, due to their training on human feedback. 
In social science, LLMs are used for tasks related to social science research questions \cite{dey-etal-2024-socialite}, in particular to detect complex social constructs in text \cite{weber_evaluation_2023}, including sexism \cite{sen_people_2023}, which is the focus of this paper.\footnote{\textcolor{red}{WARNING: This paper researches sexism, and includes sensitive and hateful content. The researchers in no way condone sexism or hate of any kind.}}

LLMs can be evaluated with computational measurements, e.g. by testing the performance of a model on datasets that have been annotated for a specific construct. 
However, this approach is not the only form of evaluation, and it can miss important nuances that knowledgeable domain experts want to consider when addressing their research question using these models. 
Combining the strengths of such nuances in human expertise with the strengths of computational models is called \textit{hybrid intelligence} \cite{dellermann2019hybrid}. This means that each (human and model) perform better at complex tasks together than they could do separately. 

Our research pipeline analyzes this connection between human expertise and LLMs. We address two high-level questions:\newline
\noindent\textbf{Q1} How do experts interact with instruction-tuned LLMs when assessing them for further use in research? \newline
\noindent\textbf{Q2} What do we learn from the expert / LLM interactions, and can we use these insights for improving computational detection? 

We focus on \textbf{sexism detection} to answer these questions, and do so by analyzing interactions between sexism researchers and LLMs in a four-component pipeline. These four study components are explained below, and illustrated in Figure~\ref{fig:process}.


In \textbf{Part I: Survey}, 9 expert participants answer questions about their expertise in sexism research, as well as their habits when using and evaluating LLMs. \textbf{Part II: Interaction} then has these experts use an interactive interface to observe and record their interactions with LLMs. First, sexism experts are asked to assess the model for suitability for sexism research in any way they see fit. We then conduct a qualitative analysis of the interactions by creating a taxonomy of strategies. Some experts ask for definitions, others generate examples, or let the LLM analyze examples. 

After that, \textbf{Part III: Construct Definition Co-Creation} asks experts to co-create sexism definitions with the LLMs. For each expert, we collect three definitions: one produced before interacting with the LLM, i.e., their own working definition of sexism (expert-written), the one that they deemed best among those produced by the LLM (LLM-generated), and the one that the expert co-created with the LLM, i.e., by editing and adding new aspects to the definition. 

Lastly, \textbf{Part IV: Modeling} uses these definitions in zero-shot sexism classification. The three definitions (expert-written, LLM-generated, co-created) from the nine experts are used in a prompt template to detect sexism with zero-shot classification in 2.500 texts from five sexism benchmarks, using LLM \texttt{GPT4o}. In these 67,500 classification decisions, we find that expert-written definitions perform poorly, 
while models perform better with LLM-generated definitions, and 
only some experts improve performance with co-written definitions. 

Our contributions are at multiple levels:\footnote{All experimental artifacts, including survey templates, code, data, and other material are available in our repository: \href{https://github.com/myrthereuver/ExpertInteractionsZeroShotSexism}
{github.com/myrthereuver/ExpertInteractionsZeroShotSexism}}

\begin{itemize}[leftmargin=*, itemsep=0.2pt, topsep=0pt]
   \item At the \textit{methodological level},  our study is the first to combine the different methodologies (survey, interactive experiment, modeling)
in one pipeline (refer to Section~\ref{sec:relwork} for a detailed discussion of previous work). Our contribution is a framework and method for eliciting and recording multi-turn LLM-human interactions, specifically for researching collaboration and hybrid intelligence for construct detection; 
    \item At the \textit{level of novel resources}, our contribution is a  dataset containing a) expert-LLM interactions on sexism b) sexism experts' ratings of \texttt{GPT4o}'s suitability for sexism research, as well as c) for each expert, three different definitions of sexism (an expert-written, LLM-generated, and co-created one). These datasets are complemented by the insights we gathered through in-depth qualitative analysis of the interactions between human experts and LLMs. 
    \item At the level of \textit{computational modeling}, we use the collected definitions and employ them for zero-shot LLM classification, effectively connecting findings on the expert's expertise, prompting strategies, and model performance on sexism detection, as well as researching the impact of hybrid intelligence on zero-shot classification. 
\end{itemize}


\section{Related Work}
\label{sec:relwork}

Our work relates to several domains of research: interaction between humans and language technology, zero-shot prompting for complex construct detection, and the use of definitions in prompting. This section identifies how our study fills a research gap at the connection between these domains.

\paragraph{How People Use Language Technology} 

Earlier work has used surveys and interviews to determine how end-users of NLP technologies conceptualize and use these systems~\cite{10.1145/3544548.3581196}. This is especially common in cases where there is potential harm for at-risk communities, for instance with translation systems and LGTBQ+ individuals \cite{lauscher-etal-2023-em, ungless-etal-2023-stereotypes}. This is also done when end-users do not belong to at-risk communities: \citet{ter_hoeve_what_2022} surveys users on why and how they use summarization systems.

In addition to asking end-users questions, the methodology employed in this paper also belongs to the domain of ethnographic methods. Such methods aim to study interactions of people and their surroundings, and observe their behavior in natural, open-ended, and unguided settings \cite{brewer2000ethnography}. These observations are then studied to find patterns, commonalities and perhaps avenues for further experiments and research. Ethnographic analysis of language technology systems can shed light on how such systems are currently used, as well as on how the use and design of a system can be improved for contexts \cite{DesignEthnography}. Such studies for instance observe interactions with chat systems by hospital personnel \cite{wang2020please}, use of news recommendation systems \cite{schjott2023designing} or e-Commerce platforms \cite{ethnography_foodplatform}. In machine translation, \citet{desilets2008reliable} use ethnographic techniques such as contextual inquiry to analyze usage of machine translation by professional translators. Their study highlights how such observations can be used to develop new technical approaches. 

\citet{zamfirescu2023johnny} analyze the prompt designs of 10 participants with no LLM experience in a no-code environment, tasking them improve a virtual cooking chatbot. They find these non-experts lack systematicity in their prompt design, and stop easily because of errors. 

Generative models are also studied with ethnographic or observational analysis in the social sciences \cite{liu_analyzing_2023, de2024synthetic}. These studies analyze the inputs and outputs of models, and how humans react to them in the form of perceived usefulness or emotions their responses evoke in a user. However, these studies do not directly connect their observations to more traditional methods of evaluation in NLP.

\paragraph{Zero-Shot Prompting for Social Constructs}

\citet{jacobs2021measurement} introduce practices on measurement modeling from the social sciences to computer science. They argue that the computational operationalization of a complex construct should not only be evaluated on \textit{predictive validity} (e.g.\ classifying an unseen test set), but should also involve testing the broader notion of \textit{construct validity}. \citet{reuver_no_2021} show how social science theory can help looking beyond such task-based evaluation when there is a connection to a societal challenge, such as a lack of diversity in news recommendation.


Work on LLMs understanding of  constructs such as hate and racism has used zero-shot prompting, that is: asking an LLM in a language-based prompt without performing any additional training, to perform a specific classification task. For instance, \citet{shaikh_second_2023} evaluate the Chain of Thought (CoT) prompting \cite{wei2022chain} technique for identifying and responding to harmful or toxic questions about people, and find it increases both nonsensical reasoning and biased answers based on generalizations of socio-demographic aspects, e.g. "[racial group X] is dumber". 
A crucial role is here also for the nuance and human context in the prompt: \citet{beck_sensitivity_2024} find adding socio-demographic information in prompts for subjective NLP tasks can influence performance. \citet{jiang-etal-2024-examining} infuse prompts with annotator information with five prompting strategies for sexism detection with LLMs, and find that models are biased by annotators' attitudes. Therefore, information in the prompt matters. Which leads us to our next component: definitions in zero-shot prompting.

\paragraph{Definitions in Prompting for Social Constructs}

Other work also researches the use of definitions in prompts for zero-shot classification with LLMs. This is done for detecting social science concepts, and also for hate and sexism specifically. \citet{peskine_definitions_2023} research expert-written vs GPT-generated definitions for classifying tweets into different categories of conspiracy theories. Their results indicate that human-written definitions are better than ones written by LLMs, but they have no examples that are co-written between experts and LLMs. They find that GPT definitions similar to human-written definitions are better in performing on unseen test sets. 
\citet{khurana-etal-2025-defverify}, building on earlier work that looks into different granular aspects of hate speech definitions \cite{khurana-etal-2022-hate}, analyze whether Transformer models actually reflect their dataset's definitions of hate speech. It introduces a method using a user-specific definition of hate speech, and quantifies to what extent a model reflects the intended definition. They find most models do not capture the aspects of hate that are defined in their dataset's definition. 
Most recently, \citet{korre2025untangling} present a dataset of hate speech definitions and analyzes the semantic properties of definitions and their classification performance in zero-shot classification with LLMs. They find that hate speech definitions and their components are culturally specific. 

\paragraph{Research Gap} Earlier work has observed or interviewed end-users of LLMs and related technologies, conducted zero-shot prompting experiments with LLMs for complex construct detection, or analyzed definitions used in computational detection of sexism or hate. Others have studied the role of different definitions in zero-shot prompting, but these either use existing definitions of the construct \cite{korre2025untangling}, or compare expert-made and LLM-made definitions of the construct \cite{peskine_definitions_2023}. To our knowledge, no previous work has expert participants completing a specific task together with the LLM (interactions about sexism and co-creation of definitions of sexism), and then using the outputs of this task for zero-shot learning on benchmark datasets. This study connects these aspects in a four-component pipeline. 


\section{Methods}\label{sec:method}

Our methodology connects a qualitative, in-depth analysis of experts interacting with LLMs to computational detection of sexism, in a four-part pipeline. 
The following sections describe every step in this pipeline.

\subsection{Part I: Pre-Interaction Survey}\label{sec:survey}

First, we conduct a pre-interaction survey on our participants, who are sexism experts. Our intention with the survey step is twofold: (1) find how and how confidently researchers of sexism use LLMs, and (2) collect information on the participants for analyzing their interactions later in the pipeline.

\paragraph{Participants}
We reached out to sexism experts within and outside our network. Study participation took place from June until September 2024. Initially, 11 experts completed the pre-interaction survey; however, two participants (experts 7 and 8) did not complete the interactive experiments, which resulted in nine fully completed experimental responses. A participant dropout rate of 18\% is considered a low dropout rate for online experiments, with previous work mentioning 20\% \cite{gagne2023run} to 30\% \cite{galesic2006dropouts}. 

Our participants are sexism researchers connected to research universities and institutes in Europe and the United States, with varying levels of computational experience. Nine participants is a small group, but is considered a suitable number of participants for a qualitative, in-depth study \cite{guest2006many, hennink2022sample}. Participants gave their informed consent about the study and its purpose. Participants were offered 6 euros for 30 minutes of participation, the minimum wage in Germany, which some declined.\footnote{We completed an ethics check about our study at the VU Amsterdam, and participants were explicitly debriefed about the potential of seeing harmful responses from the LLM. See Appendix~\ref{erb-check} for more information on our precautions to ensure a responsible participant study.}

All participants are self-identified researchers of sexism, with a mean of 6.09 (SD = 2.9) years of expertise in research related to sexism, ranging in career stage from doctoral student to assistant professor. Only one participant was a man, the rest identified as women.\footnote{More detailed information about the participants can be found in Appendix~\ref{app:survey}.}

\paragraph{Survey Design}

Our survey was designed with Qualtrics,\footnote{\url{https://www.qualtrics.com/}} with responses recorded anonymously. In addition to information about their expertise and career stage, we ask experts about their previous experience with LLMs. 
We then use four items from a 1-7 Likert scale to assess participants' self-confidence in sexism research. These were adapted items from a validated expertise and self-confidence measurement tool called the Collective Self Esteem Scale (CSEM) \cite{luhtanen1992collective}, where we used items related to professional self-esteem. Validation means the scale has been tested to consistently and accuracy measure confidence \cite{boateng2018best}. This scale contains items like "I have done substantial research related to sexism, hate towards women, or related concepts" and "I feel like I have substantially less ability in detecting or researching sexism than others." 

Our second set of items elicited attitudes towards LLMs in social science research. Participants responded on a 1 - 5 agreement Likert scale to statements such as "I want to use LLMs in my next research project" and "I want to learn more about LLMs". These questions were not taken from a standardized measurement instrument, and we used a standard 5-point Likert scale. 
After completing this pre-survey, participants were introduced to the first interaction with the LLM.

\subsection{Part II: LLM-Expert Interactions}\label{sec:interactive}

The interactions between LLMs and sexism researchers allow us to analyze how experts assess the knowledge of LLMs about their domain of expertise. The interactions also allow us to observe experts' unguided interactions with LLMs. 

\begin{figure}[ht!]
    \includegraphics[scale=0.13]{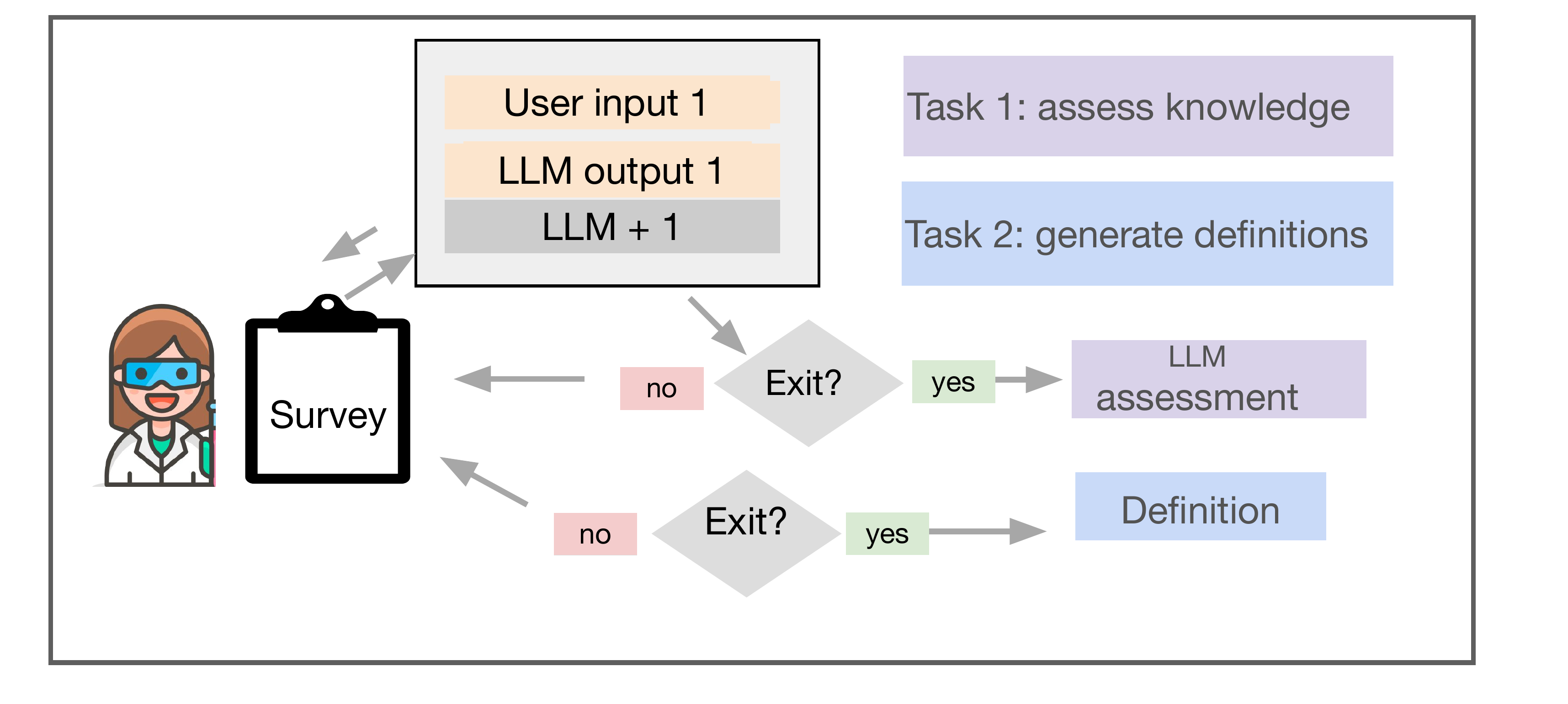}
    \caption{Explanation of the interactive experiments of Part II and Part III of our pipeline.}
    \label{fig:API}
\end{figure}

\paragraph{Experimental Design}
The participants were instructed to assess the LLM's knowledge of sexism in any way they saw fit, for a maximum of 10 input-output interactions, in an interactive environment. Our interaction model is \texttt{gpt-3.5-turbo}\footnote{Version with knowledge cut-off September 2021}. Experts could choose to end the interaction at any time by exiting a loop after indicating that they were ready to assess the model's knowledge. See Figure~\ref{fig:API} for the basic representation of the interface and loop. We release our Qualtrics template for future use of this methodology. More information on this design is in Appendix~\ref{app:Qualtrics_design}.

After the interactions, experts rate how satisfied they are with the model's knowledge and suitability for sexism detection on four items with a Likert scale range from 1 to 5 indicating from full agreement to full disagreement with the statement, which included "This model can distinguish nuances of sexism" and "I trust this model's capabilities". 




\subsection{Part III: Co-Creation of  Definitions}\label{sec:Co-Creation}

\paragraph{Experimental Design}
Our second interactive experiment concerns co-creating definitions of sexism. The participants are told that the definitions will be used for sexism detection. They are then asked for their own comprehensive definition of sexism (to which we will refer in the reminder of the paper as \textit{expert-written}). The participants also have available a maximum of 10 interactions (prompts) to co-create a definition with the same model. In this set-up, the participants can again choose to end an interaction any time. At the end of the interaction, the participants are asked to review the full interaction and perform two actions: a) select their preferred definition among those generated by the LLMs (referred to in the paper as \textit{LLM-generated} definition) and b) copy and edit their preferred definition (we refer to this definition as \textit{co-created}). 
After the interactions, the participants were asked to complete a 7-point Likert scale to rate their satisfaction with the co-created definition, as well as rate on five different aspects of the definition: validity, comprehensiveness, simplicity, covering all aspects of sexism, and its depth. Afterwards, participants could exit the survey.  

\subsection{Part IV: Modeling}\label{sec:modelling}


Our modeling experiments consist of zero-shot classification with generative LLMs for sexism detection. These experiments connect the experts' strategies and definitions to benchmark performance, and measure the impact of \textit{hybrid intelligence}  \cite{dellermann2019hybrid} - whether co-creating definitions allows models to use the strength of both expert and model knowledge. Although LLM classification has limitations, in particular due to the reliability and reproducibility issues of these models \citep{reiss2023testing}, we still consider it a relevant benchmark due to the growing application of LLM and the potential for applied measurement designs~\cite{atreja2024prompt}.


\paragraph{Definitions and Prompts}
Our modeling experiments use three definitions of sexism from each of the nine experts: the \textit{LLM-generated} definition, the \textit{expert-written} definition, and the \textit{co-created} definition. This leads to 27 different definitions of sexism used for prompting. 
Our prompt template was based on the one in \citet{sen_people_2023} (see Appendix~\ref{app:model_specs}), and completed with each of the 27 definitions of sexism.\footnote{Table~\ref{tab:alldefinitions} in the Appendix contains all these definitions.} The LLMs were prompted to respond with "sexist", "not sexist", or "don't know". 

\paragraph{Models}  We use \texttt{GPT4o}\footnote{\url{https://openai.com/index/hello-gpt-4o/.}} for our zero-shot prompting for sexism detection. 
We report generated results with a temperature of 0, since this setting is the most deterministic, and a higher temperature leads to generation of less probable tokens \cite{renze-2024-effect}. A temperature of 0 is commonly used by social scientists when using ChatGPT for labelling social constructs \cite{10367969}, and also by recent computational work assessing the effect of different definitions on zero-shot LLM prompting \cite{korre2025untangling}.\footnote{We completed additional experiments with a different GPT temperature and a second LLM, \texttt{LlaMa-3.1-70B-Instruct} \cite{dubey2024llama}. Our prompting with LlaMa, see Appendix \ref{LLama_results}, shows lower results and less variability over definitions and datasets. The same counts for a higher temperature of GPT4o, see Appendix~\ref{app:temperature}.}


\paragraph{Evaluation}
To limit the computational costs\footnote{See Appendix~\ref{app:cost} for a specification of costs.} while still ensuring coverage of different sexism benchmarks, we use samples rather than full benchmark datasets for zero-shot classification. We evaluate on 500 annotated texts randomly sampled from each of the following English-language sexism datasets. These contain content from several social media platforms or content resembling social media text:

\begin{enumerate}[noitemsep]
    \item The ``Call me Sexist But'' (``CallMeSexist'') dataset collected by \citep{samory_call_2021}, which has three types of data --- tweets, survey scale items assessing sexist attitudes, and adversarial or counterfactually augmented versions of the two former data types.
    \item ``Explainable Detection of Sexism'' (EDOS) dataset, which is based on a shared task in Semeval 2023, with sexist texts from Reddit and Gab~\cite{kirk2023semeval}.
    \item ``Reddit Misogyny dataset'' curated by \citet{guest2021expert}.
    \item The EXIST sexism dataset, comprising of tweets~\cite{EXIST2021}
    \item Finally, we also use a subset of the Hatecheck dataset~\cite{rottger2020hatecheck}, where the target of hate is women. This dataset consists of test suites for evaluating the robustness of hate speech classifiers. 
\end{enumerate}

Several of these datasets have fine-grained sexism categories \cite{samory_call_2021,kirk2023semeval}. However, these categories are neither consistent across datasets and nor do they map to the dimensions invoked in the expert and LLM definitions, hence we use the binary sexism labels.

In terms of class distribution, our samples are representative of the original datasets. The RedditGuest dataset (13\% sexism) has sexism as a rare class compared to the CallMeSexist and Hatecheck datasets (closer to 50\% each class). The EDOS and EXIST datasets are somewhat in between these, with each around 25\% sexist. We believe these samples a) fairly represent performance on the respective benchmarks and b) fairly encode that sexism vs. non-sexism content in the real-world online social media contexts that these datasets represent, where sexism is not always 50\% of the data. See Appendix~\ref{app:data_stats} for the respective distributions per full and sampled dataset.



\section{Results}\label{sec:results}

\subsection{Part I: Pre-survey}

Below we discuss the key results of our pre-survey measurements. Detailed results for all experts and outcomes are in Appendix~\ref{app:results_survey}.

\paragraph{Attitudes towards LLMs}
Experts' attitude towards LLMs, and how confident they are in using and evaluating them, are moderately positive to high (M = 3.47, SD = 1.06) for all experts, with Expert 6 the only negative outlier. Notably, this is an expert without direct prior LLM knowledge.

\paragraph{Self-confidence on sexism research}
All the experts self-report to be experienced researchers of sexism (M = 5.55, SD = 1.23) and are confident about their own knowledge of sexism (M = 5.91, SD = .67).\footnote{The difference in these two averages is due to Expert 10 reporting a low score (2) on the first question, however they also reported a total of 8 years experience in the social science research, and given their answers to other items, we hypothesize this was an error possibly due to the two earlier survey items being framed negatively (i.e. phrased with a negation, such that a low score actually meant high confidence).}.

\subsection{Part II: Interactions with LLMs} 

\paragraph{Interaction Data}
Out of the 10 possible interactions, participants use an average of 4 interactions (min: 1 - max: 7) with a standard deviation of 1.63. Broadly, experts use different strategies, from asking questions to classification examples. 

\paragraph{Model Suitability Rating}
After interacting with the LLM, participants rate its suitability for research on sexism moderately high on a 5-point Likert scale
M = 3.47 (SD = 1.06).

Expert 6, who previously reported low confidence in using LLMs, reports the second highest suitability (M = 4.75).

\paragraph{Qualitative Coding and Analysis of Interactions}\label{subsec:Knowledge_interact_analys}

We follow an inductive approach to create a taxonomy of different types of interactions between experts and the LLMs. We create two disjoint, but related, taxonomies --- one characterizing the experts' questions, directions, and instructions to the LLMs, and another for characterizing the LLMs' responses. For both, we use a grounded theory-based approach~\cite{charmaz2015grounded}, where two annotators (both of whom are authors of the papers) independently assess the prompts (for expert taxonomy) and the responses (for the LLM response taxonomy), coming up with categories. The annotators then discuss to create a taxonomy consisting of shared categories, merging certain categories, and resolving disagreements. The taxonomy is then applied to the LLM-expert interactions. Next, a third annotator (also a paper author) applies the taxonomy to a subset of LLM-expert interactions. 
This data-driven  categorizations of expert prompts and LLM responses is summarized in Table~\ref{tab:expert_taxonomy}.

\begin{table*}[ht]
\footnotesize
\centering
\begin{tabular}{@{}lll@{}}
\toprule
\textbf{Strategy}                & \textbf{Definition}                                                                                                                                                         & \textbf{Example}                                                                                                                                    \\ \midrule
Labeling Examples       & \begin{tabular}[c]{@{}p{7cm}@{}}Asking the LLM to label potentially sexist\\ content that has been supplied by the expert\end{tabular}                                  & \begin{tabular}[c]{@{}p{5cm}@{}}"Sentence: "Some people think that women are \\ inferior." Is the sentence sexist?"\end{tabular}                \\
Asking Questions        & Asking the LLM questions related to sexism                                                                                                                         & \begin{tabular} [c]{@{}p{5cm}@{}} "Would you say that women who are competent also appear cold/unsympathetic?"\end{tabular}                     \\
Implicit Tests          & \begin{tabular}[c]{@{}p{7cm}@{}}Assessing if the LLM displays sexist behaviors in other tasks, such as translation or content analysis\end{tabular}                   & 
\begin{tabular}[c]{@{}p{5cm}@{}}
"Translate the following to Spanish: The doctor asked the nurse to help her with the operation."\end{tabular} \\

Content Generation      & \begin{tabular}[c]{@{}p{7cm}@{}}Asking the LLM to generate sexist examples or content, either from scratch or by rewriting an expert supplied instance\end{tabular} & \begin{tabular}[c]{@{}p{5cm}@{}}  "Can you generate sexist content?"       \end{tabular}                                                                                               \\
Asking Explanations & \begin{tabular}[c]{@{}p{7cm}@{}}Ask the LLM to explain potentially sexist content\\ supplied by the expert or to explain it's own previous\\ answers\end{tabular}       & \begin{tabular}[c]{@{}p{5cm}@{}} “Women are kind and men are strong”. Can this statement be sexist? Can you explain why?"\end{tabular}        \\ \bottomrule
\end{tabular}
\caption{Qualitative analysis: a taxonomy of expert strategies to gauge LLMs' knowledge of sexism (Part II).}
\label{tab:expert_taxonomy}
\end{table*}

\begin{table*}[]
\footnotesize
\centering
\begin{tabular}{@{}lll@{}}
\toprule
\textbf{Strategy} & \textbf{Definition}& \textbf{Example} \\ 
\midrule
Direct Question & 
\begin{tabular}[c]{@{}p{7cm}@{}} Simply asking the LLM to provide a definition of sexism \end{tabular} & 
\begin{tabular}[c]{@{}p{5cm}@{}} "What is, Define, Explain sexism" \end{tabular} \\ 

Persona & 
\begin{tabular}[c]{@{}p{7cm}@{}} 
Giving a persona to the LLM \\ 
\end{tabular} & 
\begin{tabular}[c]{@{}p{5cm}@{}} 
"You are an expert in understanding slight linguistic nuances..." \\ 
\end{tabular} \\

Step-by-step reasoning & 
\begin{tabular}[c]{@{}p{7cm}@{}}  \\
Asking the LLM for a step-by-step reasoning when describing or explaining something
\end{tabular} & 
\begin{tabular}[c]{@{}p{5cm}@{}} 
"Define sexism step by step" 
\end{tabular} \\

Task definition & 
\begin{tabular}[c]{@{}p{7cm}@{}} 
Naming the specific task in which the definition will be employed \end{tabular} & 
\begin{tabular}[c]{@{}p{5cm}@{}} 
"I want to use a LLM to detect sexism..." 
\end{tabular} \\

Content Generation & 
\begin{tabular}[c]{@{}p{7cm}@{}} Asking the LLM to generate examples of specific form of sexism: subtle, edge, hostile vs. benevolent, ...  \end{tabular} & 
\begin{tabular}[c]{@{}p{5cm}@{}}"... giving examples of everyday (e.g. workplace) as well as online sexism."\end{tabular} \\

Reasoning  & 
\begin{tabular}[c]{@{}p{7cm}@{}} Forcing the LLM into a dialectic (or socratic) reasoning with a back and forth of multiple prompts \end{tabular} & 
\begin{tabular}[c]{@{}p{5cm}@{}}"What is a woman? … What is an adult female human being? …
You define women as biologically female human adults. What is a biological female?"\end{tabular} \\

Testing: side tasks  & 
\begin{tabular}[c]{@{}p{7cm}@{}} Asking the LLM to define other (similar) construct and tell the difference, or to classify comment and rewrite it in a non-sexist way\end{tabular} & 
\begin{tabular}[c]{@{}p{5cm}@{}}"Define sexism and misogyny. What are the differences?", "... how should the sentence be re-written to be non-sexist"\end{tabular} \\


Enhancing  & 
\begin{tabular}[c]{@{}p{7cm}@{}} Asking the LLM to rewrite the definition to enhance quality and clarity\end{tabular} & 
\begin{tabular}[c]{@{}p{5cm}@{}}"Edit your definition for better flow and clarity" \end{tabular} \\

\bottomrule

\end{tabular}
\caption{Qualitative analysis: a taxonomy of expert strategies to co-create a definition of sexism (Part III)}
\label{tab:interaction_taxonomy}
\end{table*}

 
\subsection{Part III: Co-Creation of Definitions}\label{sec:Co-Creation}



\paragraph{Interaction Data and Definition Ratings}

Recall that each expert produces three definitions in these interactions: \textit{expert-written}, \textit{LLM-generated}, and \textit{co-created}. This definition-related task shows more interactions with the LLM (M = 9.11 out of 10) compared to the previous task of questioning the LLM's knowledge of sexism: out of the nine participants, six use all (10 out of 10) available interactions.
 Overall experts rate the co-created definitions positively (M = 5.09, SD = 0.76).\footnote{Figure~\ref{fig:Happy_with_generated_def} in the Appendix reports the full results on participant satisfaction with the sexism definitions.} 
 

\paragraph{Qualitative Coding and Analysis of Interactions}
Following the same methodology illustrated in Section~\ref{subsec:Knowledge_interact_analys}, i.e., a ground theory approach, we identify the strategies of the experts when co-creating a definition of sexism with the LLM, and organize them into a taxonomy of strategies.
We notice strategies such as inducing personas ("You are an expert."), testing the LLM on closely related constructs (misogyny), or explicitly stating the experts' own goal ("I want to detect sexism"). See Table~\ref{tab:interaction_taxonomy} for a full overview of these, with examples.

 \paragraph{Definition Change and Similarity}\label{sec:definition_similarity}
Expert-written definitions are generally shorter (M = 34.44, tokens, SD = 25.24), than the LLM-generated definitions (M = 119.89, SD = 58.29) and the co-created definitions (M = 110.55, SD = 56.44). Most experts do minimal edits on the LLM definition for their co-created definition.

Expert-written definitions align with different aspects in definitions of the five sexism benchmarks. For example, Expert 2 wrote a definition that mentions stereotypes and disrespect, similar to the definition of EXIST data set, which also mentions stereotyping and prejudice.\footnote{Additional similarity experiments are in Appendix~\ref{app:def_analysis}.} Expert 11's initial definition is relatively short and generic, and becomes much more specific to gender issues after interacting with the LLM.

\subsection{Part IV: Modeling 
}

\begin{figure*}[ht]
    \includegraphics[scale=0.49]{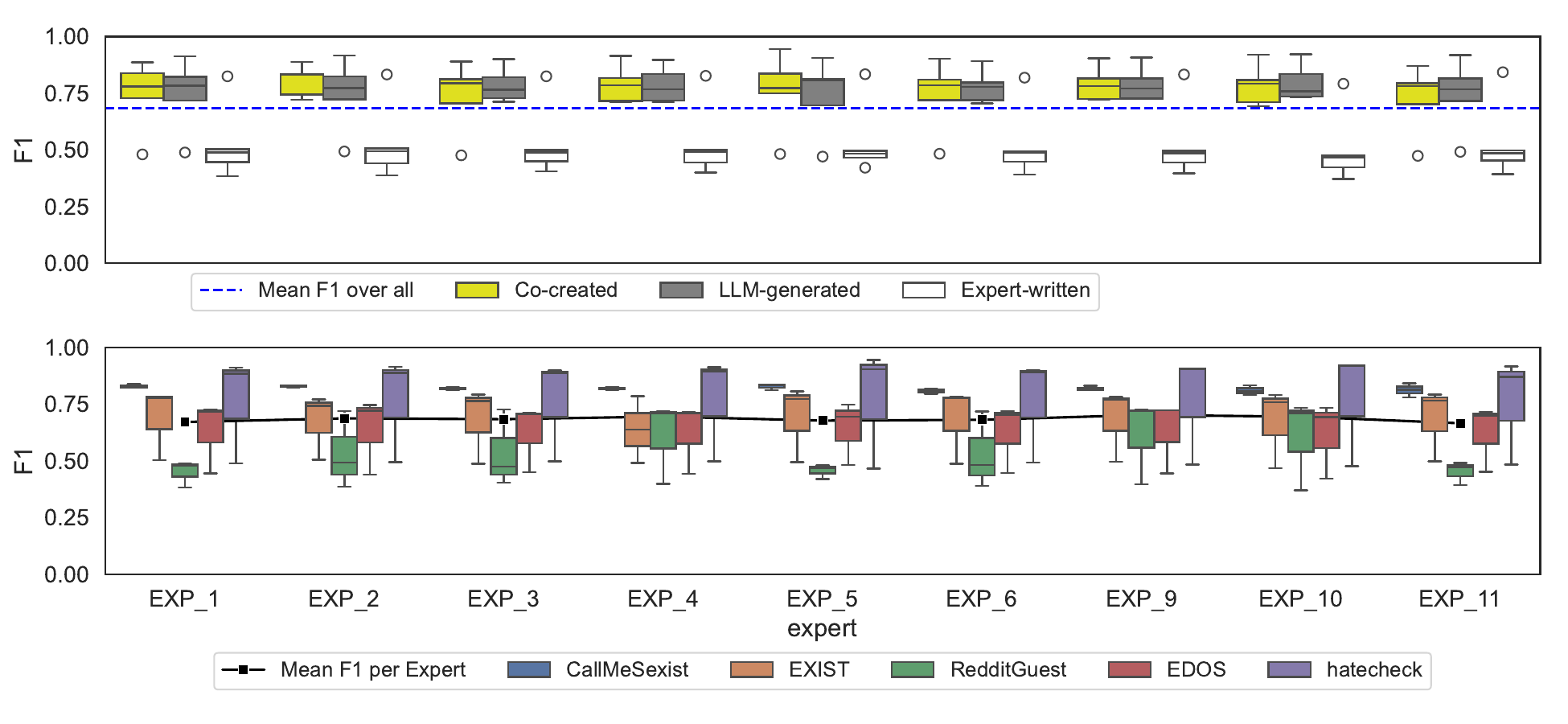}
    \caption{$F1$ (macro) performance of \texttt{GPT4o} per participant over the three definitions (upper plot) and over the five datasets (bottom row). 
    }
    \label{fig:Diff_exp}
\end{figure*}

Our modeling results answer questions about the connection between definition, expert strategy, and benchmark performance. Results consist of 67.500 classification decisions on three dimensions (three different definition types for nine different experts, on five sexism datasets). 
We therefore present mean performance over definition types, datasets, and experts in-text, and a visual overview of our results in Figure~\ref{fig:Diff_exp}.\footnote{Detailed results (per dataset and individual participant) are available in our code repository.} All results are reported in $F1$ (macro), unless stated otherwise. 

 \paragraph{Dataset dependence and majority class}
Modeling results are dataset-dependent. As seen in Figure~\ref{fig:Diff_exp}, some datasets receive zero-shot performances consistently above majority class $F1$ for all experts and definition versions, such as the CallMeSexist dataset - which also has a near 50-50 distribution of sexist vs non-sexist content. In contrast, the RedditGuest dataset responds very differently to prompts by different participants, and sometimes performs lower than majority class $F1$. This is partly explainable by the majority class being very dominant in this dataset, at 90/10 distribution, making the majority class a very strong baseline. Further analyses of experts and definitions that perform above or below a majority baseline for each of the five datasets can be found in the Appendix~\ref{app:majority_class}.

\subsubsection{Performance across Definition Type}

\paragraph{Performance over definition types}
\texttt{GPT4o} performs, over all datasets, better with the LLM-generated definitions (Mean $F1$ = .765, SD = .113) than with the co-created definitions (Mean $F1$ = .762, SD = .122), and substantially better than with expert-written definitions (Mean $F1$ = .532, SD = .154). 

\paragraph{Performance over definition types per dataset}
\texttt{GPT4o}'s performance fluctuates per dataset. On the CallMeSexist dataset, all three definition types perform nearly identical (LLM-generated definitions with a mean of $F1$ =.814, co-created definitions with a mean of $F1$ = .823, and participant definitions with a mean of .826), while the results on the RedditGuest dataset shows a considerable gap between the three definition types (LLM-generated definitions with a mean of $F1$ = .620, co-created definitions with a mean of .581, and participant definitions with a mean of $F1$ = .397).




\paragraph{Performance of definition types on "sexist" class}
There is a bigger difference in \texttt{GPT4o}'s performance when looking at the performance of the sexist vs non-sexist class. Co-created definitions have a small gain in performance over the LLM-definitions on the sexist class: LLM-generated definitions on average perform at $F1$ = .738, co-created definitions with a mean of $F1$ .740, and expert-written with a mean of $F1$ = .501).







\subsection{Difference over experts}

The difference between different experts is small: the expert with the lowest mean performing definitions is Expert 1 ($F1$ = .671, SD = .184), and the Expert with the highest mean over definitions is expert 9 ($F1$ = .702, SD = .166).

The upper half of Figure~\ref{fig:Diff_exp} shows that some experts are able to define more effective definitions of sexism than others, while others succeed better in the co-creation. Expert 5 shows comparatively higher scores for the co-creation with the LLM. This is notable because on average the LLM-generated definitions perform better.

The difference between experts becomes more pronounced when the results are disaggregated by dataset. The lower half of Figure~\ref{fig:Diff_exp} shows that CallMeSexist dataset performs > .80 in $F1$ for all experts, while the RedditGuest dataset shows great fluctuation depending on which expert has provided the definitions, with Expert 4 managing results around $F1$ = .60 while Expert 11 remains at $F1$ = .50.

\paragraph{Connection between the parts of the study}
This paper consists of four components: a survey, two interactive experiments, and zero-shot classification. When connecting results from all parts in the pipeline, we obtain additional insights. For instance, Expert 6 - less experienced with LLMs - was sometimes successful at co-creating definitions that perform better than the LLM-written definitions. In contrast, the co-created definition of experts confident in the use of LLMs (such as Expert 4) perform worse than the LLM-generated ones. This difference does not seem to be influenced by the difference in length of the definitions, or even to definition similarity to dataset definitions.

\section{Conclusion}\label{sec:discuss}

Experts have nuanced knowledge of complex constructs in their domain of expertise. Our aim was to connect this knowledge to zero-shot construct detection with LLMs, and test whether model and expert can collaborate for \textit{hybrid intelligence}: model and expert complementing one-another. We addressed two high-level questions: 

\noindent
\textbf{Q1} How do experts interact with instruction-tuned LLMs when assessing them for further use in research? \textbf{Q2} What do we learn from the expert / LLM interactions, and can we use these insights for improving computational detection?

Answering these questions led us to develop a four-part pipeline that connects LLM-expert interactions to computational detection of sexism. In \textbf{Part I}, nine sexism experts first answered questions on their use of LLMs and expertise in sexism research. \textbf{Part II} consisted of an interaction experiment, where experts assess and evaluate the knowledge of an LLM (\texttt{GPT3.5}) about their domain of expertise. These interactions also allowed us to observe experts’ interactions with LLMs. \textbf{Part III} consisted of a second expert-LLM interaction, where each of the nine experts was tasked to create three definitions of sexism: an expert-written, an LLM-generated, and a co-created definition. We then evaluated zero-shot classification in \textbf{Part IV} by prompting the LLM \texttt{GPT4o} with each of these 27 created definitions on 2,500 texts from five sexism benchmark dataset. We release the interaction framework as well as the anonymous LLM-expert interactions and the definitions for future research.

Answering \textbf{Q1}, Part II and Part III found that sexism experts use different strategies for evaluating LLMs on their domain of expertise: content generation, asking questions, and labelling examples. Most experts were moderately satisfied with the LLM's knowledge of sexism. 

On \textbf{Q2}, our modeling experiments in Part IV showed that LLM-written definitions help performance on benchmarks more than co-created definitions - which counters the hypothesis that co-creation is a fruitful manner to add expert knowledge into construct definitions for zero-shot classification. However, some experts do obtain higher zero-shot performance with co-created definitions, and confidence in LLM usage does not necessarily relate to more effective definitions: experts with low self-confidence in LLM expertise were often able to co-create more effective definitions than more LLM-savvy colleagues.

\color{black}


\newpage
\section*{Acknowledgements}
MR's contributions were funded by a GESIS Junior Visiting Researcher Grant for a fully funded research visit to GESIS in Cologne, Germany from March to April 2024. MR was until 4 December 2024 also funded by the Netherlands Organization for Scientific Research (NWO) through the \textit{Rethinking News Algorithms} project, funded via the Open Competition Digitalization Humanities \& Social Science grant (406.D1.19.073). We thank GESIS for covering the costs of proprietary models as well as payments to the participants. 

We thank our colleagues for their comments and feedback on earlier versions of this paper, especially prof. dr. Suzan Verberne and prof. dr. Antske Fokkens. The feedback provided by this paper's ARR reviewers was also excellent and enhanced this paper. All remaining errors, flaws, or unclarities are ours.

\section*{Limitations}
We identified at least three aspects that limit the generalizability of the findings in our study.

\paragraph{Scope} We only test our framework on one social construct, sexism. The results - both on the interactions and on the classifications - may therefore not generalize to other complex constructs. Additionally, we tested our framework with only one LLM (\texttt{GPT4o}), and a limited set of nine experts. Future work may want to increase the scope of this work by adding more experts, constructs, or LLMs. We did complete additional zero-shot classification experiments with different temperature settings of \texttt{GPT4o}, as well as a different LLM (\texttt{LlaMa-3.1-70B-Instruct}), which showed lower performance and less variance over differences in prompt. These analyses can be found in the appendix.

\paragraph{Representativeness of Participants} We use a limited sample of experts from WEIRD \cite{henrich2010weirdest} contexts: Western, Educated, Industrialized, Rich and Democratic. This is also relevant in a researcher context (e.g. researchers from the Global South), and limits generalizability to other research contexts. 

\paragraph{Language} Additionally, our datasets are only in the English language: both the benchmark datasets and the experiments were conducted in English, which also does not allow experts with different contexts to be included in these results.  

\section*{Ethical Considerations}

\paragraph{Harmful Content} Sexism is a great harm to society and the world. We explicitly condemn sexism, and additionally want to warn any person using these models or datasets for potentially harmful utterances in them.

\paragraph{Experimental Safety of Participants} Our participants were advised that they were free to leave the conversation at any time they felt the need. A large harm reduction factor was that our participants are experts already working on sexism, and therefore not as unprepared as crowd workers or other non-specialists to see sexist or hateful content. We caution researchers wanting to use a similar framework when participants are non-specialists, i.e. crowdworkers. 

\paragraph{Writing Assistance} Overleaf's integrated language model Writefull was used to assist in polishing and clarifying the language in this paper. This assistance, in accordance with the ACL Ethics Policy and Responsible Research Checklist, was solely used for improving the language in the paper rather than for producing new content or new ideas. All final writing is ultimately done by the authors, who are responsible for it.

\paragraph{Proprietary models}
Models of the GPT family are closed: developers have not released all information relating to the development and workings of these models. The openness of LLMs is not a binary variable: as \citet{liesenfeld2023opening} indicates, there are several dimensions that make a model open, from shared code and training data, licensing, and whether payment is required for access.
Despite this closedness, we still chose to use models of the GPT family because GPT models are especially popular with social science researchers researching constructs such as sexism. Additionally, the large-scale nature of our experiments and the lack of GPU access available to us made API requests more feasible for our experiments. 

However, we are aware that open models are better for the scientific community due to their more reproducible nature and also the lack of payment required. We realize that our choice for proprietary models leads us to contribute to additional attention to these models, which is not ideal. 

\newpage

\bibliography{naacl_camera_ready.bib}

\newpage

\onecolumn
\appendix
\section*{Appendix}\label{sec:appendix}

\section{Precautions for a Responsible Participant Study}\label{erb-check}
We completed an ethics form from the Social Science department of the Vrije Universiteit Amsterdam. This form is designed to decide whether a study requires additional ethical review. We received confirmation that indicated our methods were not harmful to participants, and could proceed without additional ethics review if taking into account responsible study design in the form of informed consent, participant payment, and warnings on the potential of harm.

A factor that mitigated the harm to our participants was that they were sexism research experts, who were duly informed about the purposes of this study. Participants were not directly exposed to hateful language and were informed that they could quit the experiment at any time. They were offered a payment for their time, at the minimum wage level in Germany.

\subsection{Informed Consent Statement}

\textbf{Explainer} This task is part of a scientific study about the use of AI in social science. The purpose of the task is to obtain data on how people - specifically social science experts - interact with so-called Large Language Models (LLMs).    The task  Experts are briefly asked to give information on their experience with LLMs. Then, experts are asked to interact with the model to determine whether this model is sufficient to detect the construct "sexism" in texts. Experts are then also asked to generate definitions with the model, which will be later used to improve LLM detection of sexism.    

\textbf{Compensation}  We estimate the survey takes 30 minutes. Our compensation for your time is 6 euros. This payment can be received by providing your email address after completing the survey. Your details will not be used for any other purpose, and not stored.      This study is conducted by researchers connected to GESIS - Leibniz Institute for the Social Sciences in Cologne, the Vrije Universiteit Amsterdam, and the University of Konstanz. Researchers are reachable through email: myrthe.reuver@vu.nl. Save this email address for any questions or concerns about this study.   

\textbf{Outcome}  The generated texts and data gathered in the study may be published, or made available to other researchers. Your name or identifying information will not be stored, and will not be shared with anyone.    

\textbf{Note}: Despite practices on model safety that went into model development, the LLM can potentially generate harmful output that may be upsetting or offensive. Such output is not condoned by the researchers or their institutes. You can stop interacting with the model at any time, for any reason.

\textbf{Consent statement}   

I understand the purpose of this study, and participate in this task out of my own free choice.    

I understand I can withdraw at any time, for any reason, without any consequences of any kind.    

I understand that in any report on the results of this research my identity will remain anonymous.   

I understand that I am free to contact the researchers through myrthe.reuver@vu.nl to seek further clarification and information.

\section{Survey and Experiment}\label{app:survey}

\subsection{Survey and Experimental Design}\label{app:Qualtrics_design}

We use Qualtrics as a survey platform, and connect calls to the OpenAI API through a Web Service component in the workflow. This Web Service had as input a text field presented to the participant, and collected the output that the API provided in an object that was presented in a text presentation question that was presented to the participant. Our Qualtrics templates, both in pdf and in Qualtrics format, are released through our GitHub repository.

The 10-turn conversational loop was achieved by a chain of if-clauses in the survey flow: if a participant chose to end the conversation, the participant was re-directed outside of the Web Service. If the participant indicated she was not yet done with the experimental interaction, she was instead re-referred to another Web Service field, but one that received as input the entire previous conversation. This made the conversation continuous despite requiring multiple API calls due to Qualtrics not allowing one call loop towards the same Web Service.

\subsection{Additional Participant Info}
Seven of our participants were PhD candidates or equivalent, one was a postdoc, and three participants assistant professor. When asked for their main research discipline, four experts answered with communication science, three experts answered with (computational) social science. The remaining participants responded with Human-Computer Interaction or computer science. This broad range of answers assures a wide range of experiences with LLMs for sexism research in our study.

\subsection{Additional Expert Survey results}

\textbf{LLM Experience}
Nine out of the 11 social science experts had used LLMs at least once for research purposes. Of the ones who use LLMs, one expert indicates using LLMs daily, 5 indicate a weekly use, one indicates monthly usage while two experts say they use LLMs less than monthly. All experts who used LLM were familiar with ChatGPT, with Claude and Mistral being the second-most popular LLM (both N = 4). 

All experts who had used LLMs before were familiar with interacting with an LLM through a chat interface, and eight experts were familiar with using LLMs through API or code. Fine-tuning was used before by seven experts. A minority of experts (2 experts) indicated experience with designing LLM agents and prompt chains.

Experts indicated from a list which tasks they most used LLMs for: data annotation (N = 5), followed by assistance with text writing (N = 4), followed by summarization (N = 2) and information seeking (N = 2). One experts mentioned help with data visualization. Interestingly, a connection could be seen with background expertise: experts with a background in communication science all mentioned text writing, while experts mentioning computational social science all mentioned data annotation.

In an open question about LLM use in research, expert 3 indicated feeling uncomfortable with over-use of LLMs for data annotation and expert 11 indicated to find anthropomorphization of LLMS to be a potential danger for social scientists using LLMs. Expert 6 indicated they did not use LLMs herself, but relied on co-authors to use LLMs in their shared projects. Expert 9 shared that they not only use LLMs in sexism research, but researches the use of LLMs in social science.

\subsection{Heatmaps of Questionnaire Likert Scales Results}\label{app:results_survey}

\begin{figure*}[h!]
\centering
    \includegraphics[scale=0.40]{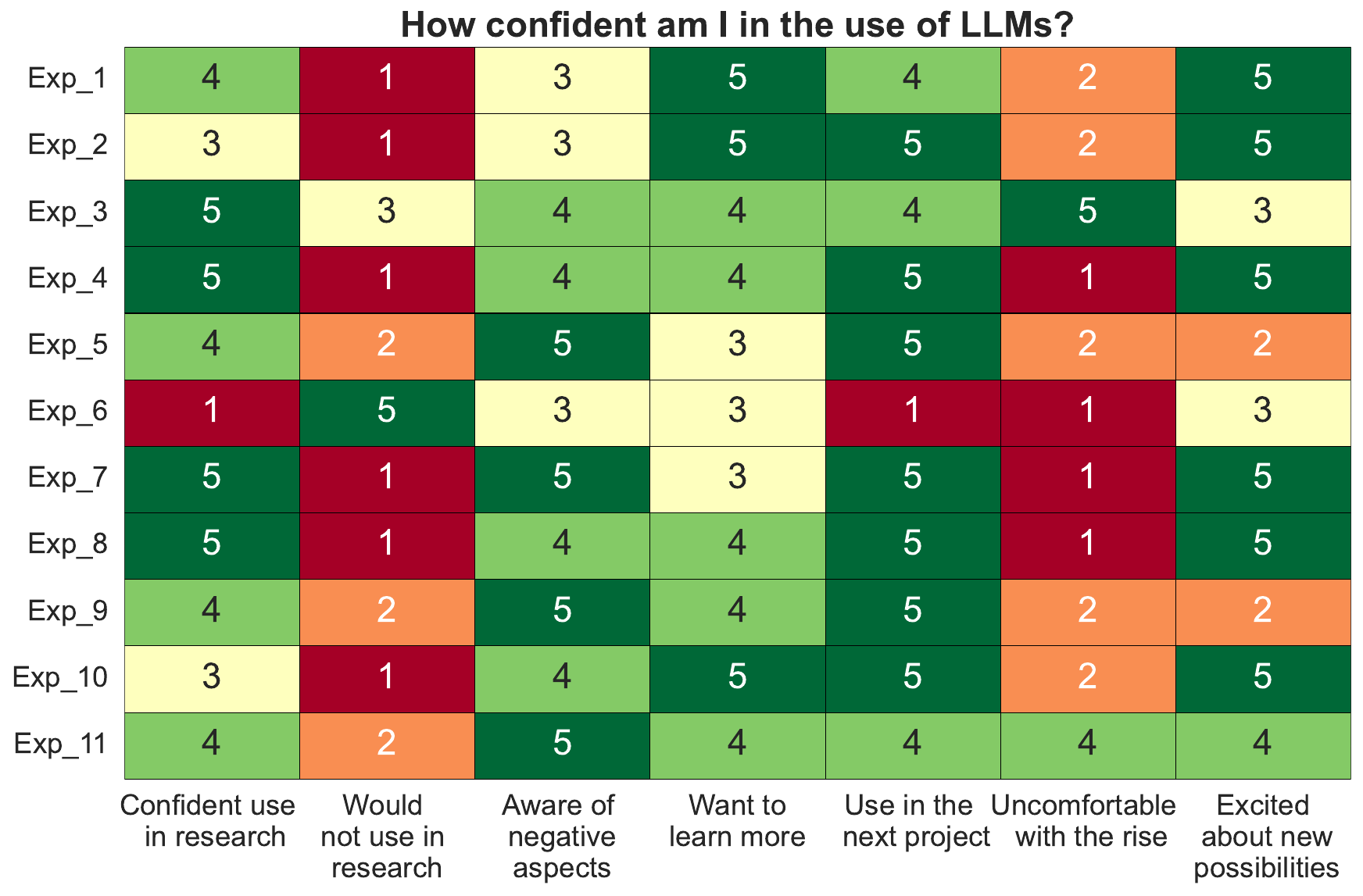}
    \caption{Heatmap of Likert scale on participants experience on LLMs.}
    \label{fig:Experience_with_LLMs}
\end{figure*}

\begin{figure*}[h!]
\centering
    \includegraphics[scale=0.40]{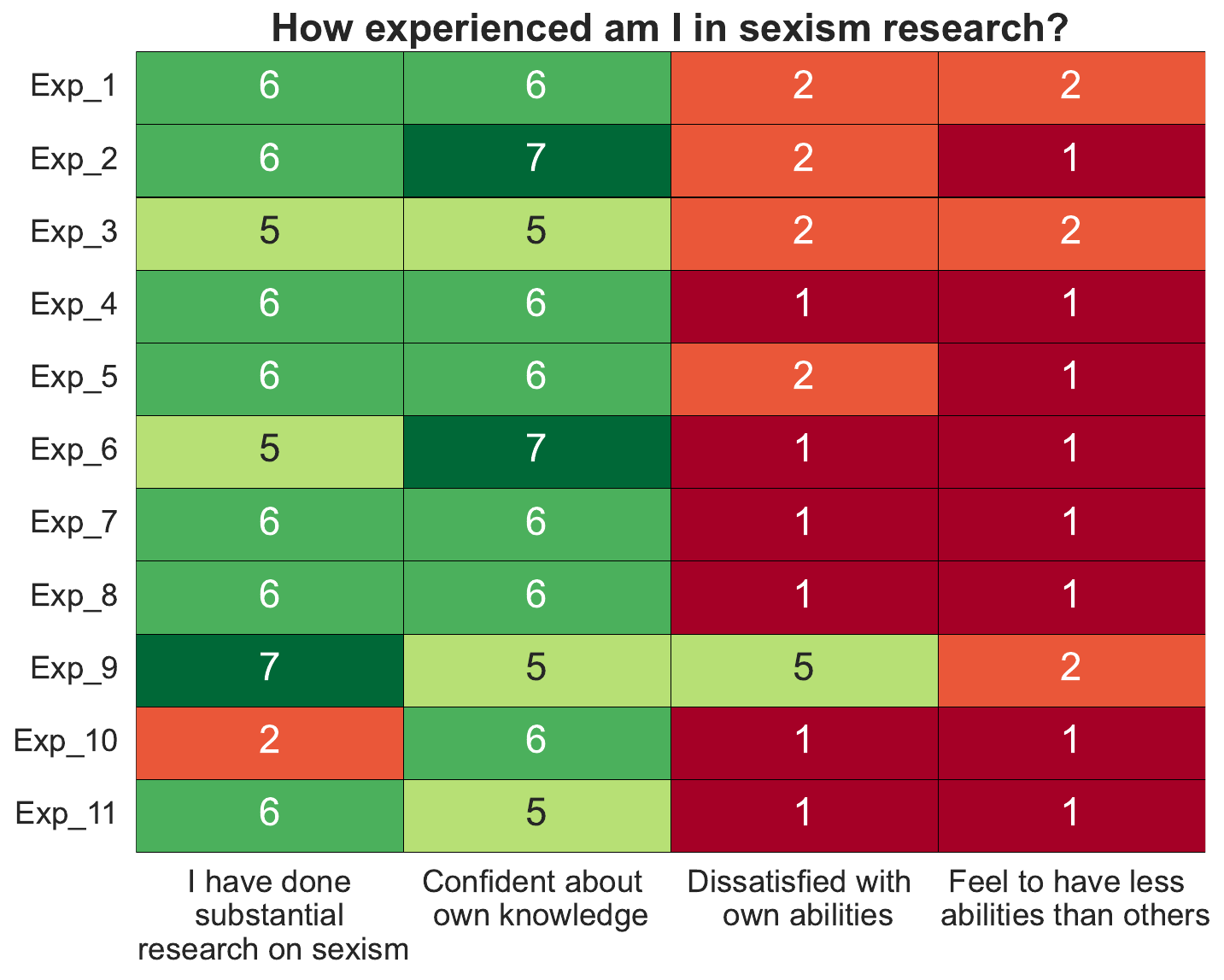}
    \caption{Heatmap of Likert scale auto-reported experience on sexism research.}
    \label{fig:experience_sexism}
\end{figure*}

\begin{figure*}[h!]
\centering
    \includegraphics[scale=0.40]{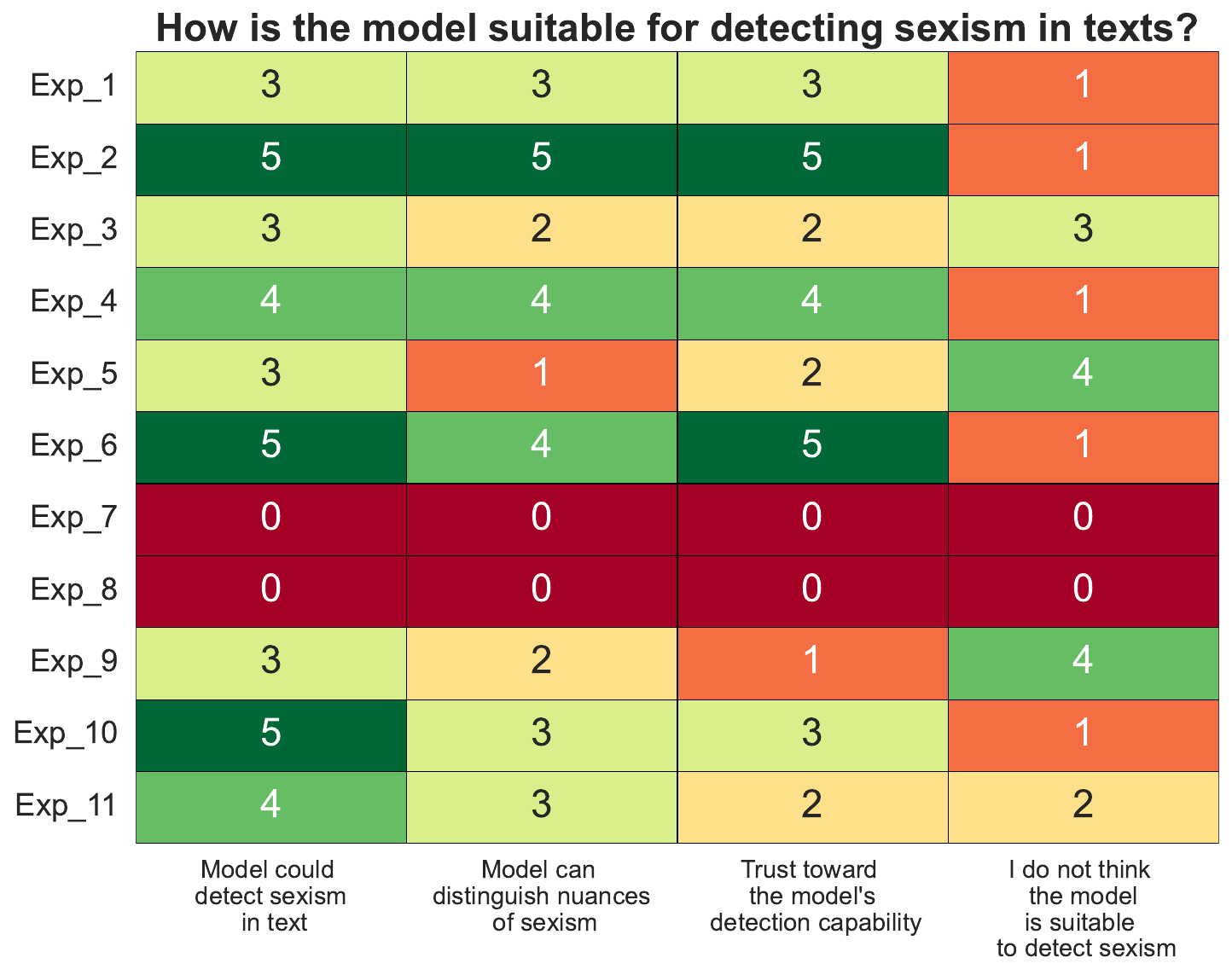}
    \caption{Heatmap of Likert scale on suitability of the model in detecting sexism.}
    \label{fig:model_suitability_for_detecting_sexism}
\end{figure*}

\begin{figure*}[h!]
\centering
    \includegraphics[scale=0.40]{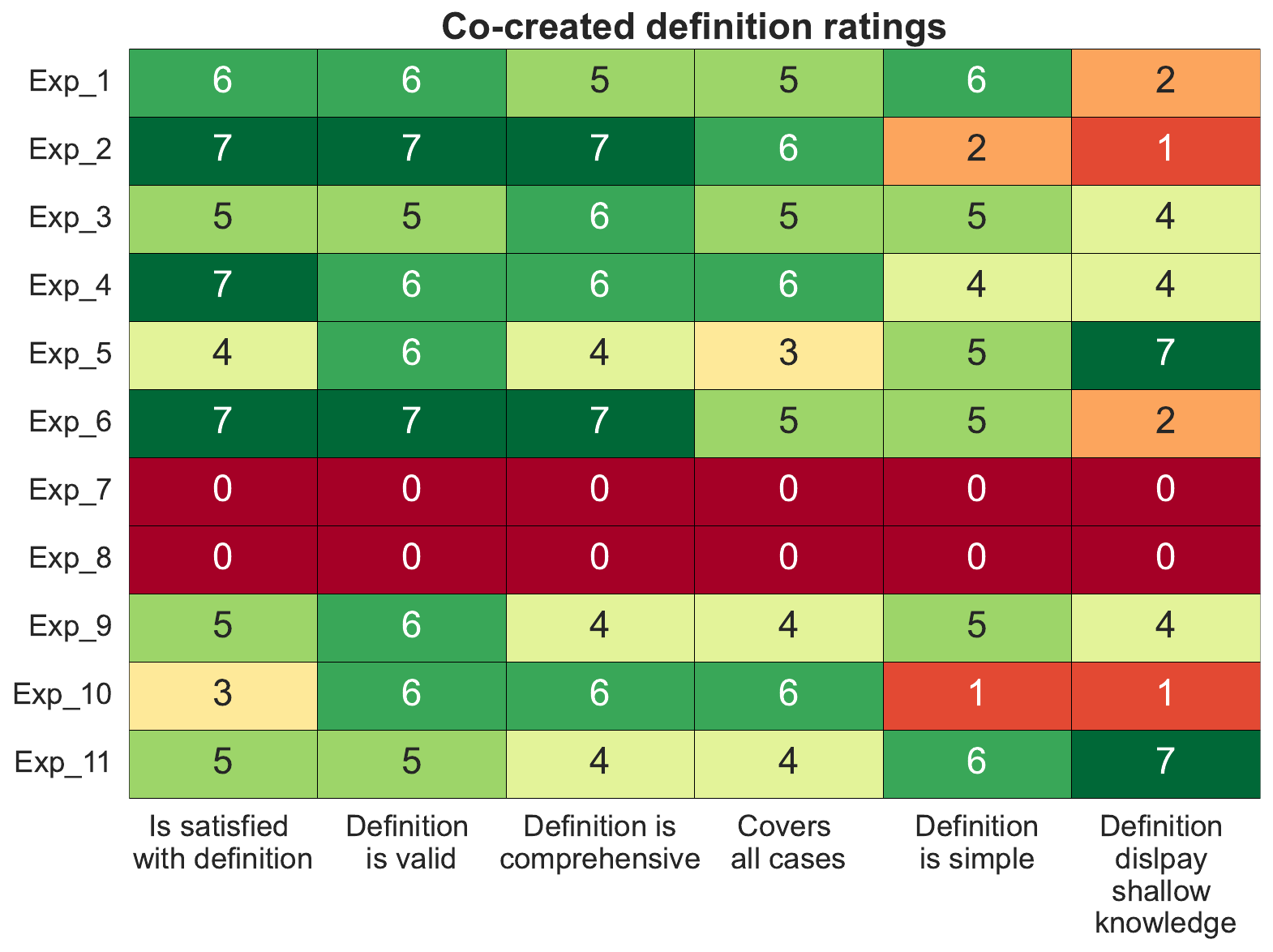}
    \caption{Heatmap of Likert scale on the suitability fo the co-created definition.}
    \label{fig:Happy_with_generated_def}
\end{figure*}

\section{Modelling Specifications}\label{app:model_specs}
Our experimental code is written in Python 3.9.5 and it uses the openai package version 1.51.2 for calling to the OpenAI API. 


\subsection{Model Hyperparameters and Compute Infrastructure}

The generation parameters are set to: temperature = 0,
max\_tokens=200, top\_p=1, frequency\_penalty=0,
presence\_penalty=0. We then repeated this with another run with a set temperature of 0.7, and max\_tokens to 20, to analyze robustness of this approach in different temperatures. Results of these experiments and an analysis of them are presented in Appendix~\ref{app:temperature}.


\newpage
\subsection{Costs for proprietary models}\label{app:cost}

Our costs for using proprietary models were paid by GESIS. 

\paragraph{Costs related to interaction experiments}

Participant interactions came at maximum 20 calls per participant, for 9 participants, at approximately 0.00317 dollar cents per call. This meant that the total costs for the experiments were around \$0.50 cents.

\paragraph{Costs related to classification experiments}\label{GPT_costs}

For the \texttt{GPT4o} classification experiments, one total run equals 9 participants times 3 different definitions, which is 27 definition prompts. These prompts were used on 2,500 texts, which resulted in 67,500 prompts sent to \texttt{GPT4o} through the OpenAI API. These 67,500 classification decisions cost approximately 0.00317 dollars per call, or \$213 in total for the entire dataset. We did these experiments twice to check for robustness, leading to a total cost of \$426.

\subsection{Prompt for classification}

"Your task is to annotate if a piece of text is sexist or not. Later, you get a detailed definition of sexism. \

Definition of sexism: \textbf{definition} 

With this definition in mind and a piece of text, please reply with either of the three options and nothing else: 1) sexist, 2) non-sexist, 3) can’t say. Please use the last option as little as possible and if you do, provide a one sentence explanation. Do not give an explanation if you pick ’sexist’ or ’non-sexist’. \

We are ready to begin now. Is this sentence sexist based on the past definition’ : \textbf{text} - ’Please reply with either of the three options and nothing else: 1) sexist, 2) non-sexist, 3) can’t say."


\section{Evaluation datasets: class balance}\label{app:data_stats}


\begin{table}[h]
\centering
\begin{tabular}{c|ccccc}
\toprule
Relative & CMSB & EDOS & REDDIT & EXIST & HateCheck \\
\midrule
Sexist        &  0.448   &  0.232 &     0.13   &    0.476   &  0.734  \\
Non sexist    &   0.552 &   0.768 &  0.87   &    0.524   &   0.266  \\
\bottomrule
\end{tabular}
\caption{Original class distribution in the five datasets - relative amount of positive and negative class}
\end{table}

\begin{table}[h]
\centering
\begin{tabular}{c|ccccc}
\toprule
In raw counts & CMSB & EDOS & REDDIT & EXIST & HateCheck \\
\midrule
Sexist        & 534 &   4854  &   699   &  1636  & 373\\
Non sexist    &  690 &  15146   &  5856 & 1800  & 136 \\
\bottomrule
\end{tabular}
\caption{Original class distribution in the five datasets - raw counts of positive and negative class}
\end{table}


\begin{table}[h]
\centering
\begin{tabular}{c|ccccc}
\toprule
Relative & CMSB & EDOS & REDDIT & EXIST & HateCheck \\
\midrule
Sexist        &  0.436 &   0.243  &   0.107   &  0.476  & 0.739 \\
Non sexist    &  0.564 &  0.757   & 0.893  & 0.524  &  0.267 \\
\bottomrule
\end{tabular}
\caption{Class distribution in the dataset subsamples we used for evaluation - relative amount of positive and negative class}
\end{table}

\begin{table}[H]
\centering
\begin{tabular}{c|ccccc}
\toprule
In raw counts & CMSB & EDOS & REDDIT & EXIST & HateCheck \\
\midrule
Sexist        & 224  &   116  &   65   &  238  & 367\\
Non sexist    & 276  &  384   & 435  &  262 & 133 \\
\bottomrule
\end{tabular}
\caption{Class distribution in the dataset subsamples we used for evaluation - raw counts of positive and negative class}
\end{table}

\section{Definitions Analysis}\label{app:def_analysis}

Table~\ref{tab:alldefinitions} lists the definitions collected from our experts in the second part of our survey. For each expert, we report the definition provided in the beginning of the experiment (\textit{Expert-written}), the definition that was chosen as the best provided by the model (\textit{LLM-generated}), as well as the co-created definition that was submitted by the expert as a response in the end of the survey (\textit{Co-created}). We also display the dataset-specific definitions employed for the annotation in the five datasets we conduct our modeling experiments on. Taken together, the definitions listed in Table~\ref{tab:alldefinitions} constitute the \textit{prompt repository} we evaluate and release in this paper.

\begin{longtable}{|p{0.1\textwidth}|>{\centering}m{0.1\textwidth}|m{0.72\textwidth}|}
\hline
\multicolumn{1}{|c|}{\textbf{Source}} & \textbf{Type} & \textbf{Definitions} \\
\hline
\endfirsthead

\hline
\multicolumn{1}{|c|}{\textbf{Source}} & \textbf{Type} & \textbf{Definitions} \\
\hline
\endhead

\hline
\endfoot

\endlastfoot

\multirow{3}{*}{Expert 1}& Expert-written & Sexism is discrimination based on gender, including stereotypes and prejudice. It can take many forms, from offensive language to sexual violence. \\
\cline{2-3}
 & Co-created  & Sexism refers to prejudice, discrimination, or stereotyping based on a person's sex or gender, typically with a bias against one gender over the other. It can manifest in various forms, including unequal treatment, unfair expectations, and harmful attitudes or beliefs about individuals based on their sex. Examples of everyday sexism in the workplace could include paying women less than men for the same job or assuming certain roles are only suitable for one gender. Online sexism could involve making derogatory comments based on gender or perpetuating harmful stereotypes in digital spaces.  \\

\cline{2-3}
 & LLM-generated & Sexism refers to prejudice, discrimination, or stereotyping based on a person's sex or gender, typically with a bias against one gender over the other. It can manifest in various forms, including unequal treatment, unfair expectations, and harmful attitudes or beliefs about individuals based on their sex. Examples of everyday sexism in the workplace could include paying women less than men for the same job or assuming certain roles are only suitable for one gender. Online sexism could involve making derogatory comments based on gender or perpetuating harmful stereotypes in digital spaces. If you have any more questions or need further clarification, feel free to ask!\\
\clearpage
\multirow{3}{*}{Expert 2} & Expert-written & I would define sexism as a way to imply stereotypical or unrespectful opinion on gender.  \\
\cline{2-3}
 & Co-created  & Sexism is a form of prejudice, stereotyping, or discrimination based on gender, often targeting women. This bias reflects the belief in one gender's superiority, resulting in unfair treatment, negative stereotypes, and limited opportunities solely due to gender. For example, a common illustration of sexism is the gender pay gap, where women are paid less than men for the same work. Sexism can manifest in different forms like unequal treatment, unfair stereotypes, or restrictions on opportunities, impacting both individuals and society. If you require more examples or details for each form of sexism, feel free to ask! Sexism's effects can be harmful, leading to inequalities, perpetuating stereotypes, and limiting opportunities for individuals. It primarily affects women but can also impact men in different contexts. Resolving sexism involves efforts at individual, societal, and institutional levels through education, promoting gender equality, implementing fair policies, and challenging discriminatory practices. If you need further clarification or details on this topic, please let me know!  \\
\cline{2-3}
 & LLM-generated & Sexism is a complex issue that involves prejudice, stereotypes, or discrimination against individuals based on their gender. This kind of bias often targets women and embodies the belief that one gender is superior to another. It can take the form of unfair treatment, harmful stereotypes, or limitations on opportunities solely because of a person's gender. Sexism is not limited to individual interactions but can also be ingrained in societal structures, leading to significant adverse effects on both individuals and society.  \\
 \hline
\hline
\multirow{3}{*}{Expert 3} & Expert-written & Believing that one sex is superior to another. It includes different forms of harmful stereotyping including benevolent sexism, etc. \\
\cline{2-3}
 & Co-created  & Sexism refers to discrimination, prejudice, or stereotyping based on a person's gender. It can manifest in various forms, such as treating individuals differently because of their gender, assuming certain characteristics or behaviors based on gender, or denying opportunities to someone because of their gender. Sexism can affect people of all genders, but it is often directed towards women and can have harmful effects on their lives and opportunities. There are different types of sexism, such as hostile sexism, benevolent sexism, and ambivalent sexism.  
 \\
\cline{2-3}
 & LLM-generated & Sexism refers to discrimination, prejudice, or stereotyping based on a person's gender. It can manifest in various forms, such as treating individuals differently because of their gender, assuming certain characteristics or behaviors based on gender, or denying opportunities to someone because of their gender. Sexism can affect people of all genders, but it is often directed towards women and can have harmful effects on their lives and opportunities. There are different types of sexism, such as hostile sexism, benevolent sexism, and ambivalent sexism. \\  
\hline
\hline

\multirow{3}{*}{Expert 4} & Expert-written & A prescriptive set of behaviors or qualities, that women (and men) are supposed to exhibit to conform to traditional gender roles.   \\
\cline{2-3}
 & Co-created  & It's fascinating to see your expertise in discerning subtle linguistic cues to identify sexism in texts. The definition of sexism centers on discrimination, prejudice, stereotyping, and biased behaviors toward individuals based on gender, leading to unequal treatment and marginalization. Evaluating texts for sexism entails examining portrayals of gender superiority or inferiority, reinforcement of stereotypes, and gender-based limitations, which involves analyzing the text's structure, tone, and phrasing to ascertain if it perpetuates gender inequality. \\
\cline{2-3}
 & LLM-generated & Sexism is a form of discrimination, prejudice, or stereotyping based on a person's gender. It involves attitudes, beliefs, and behaviors that differentiate, marginalize, or treat individuals differently based on their gender. Sexism can manifest in various ways, such as through language, actions, or systemic biases that perpetuate inequality between genders. When analyzing a text for sexism, it's important to assess if it portrays one gender as superior or inferior to the other, reinforces harmful stereotypes, or limits an individual's opportunities or worth based on their gender. This thorough evaluation considers the structure, tone, and phrasing of the text to determine if it perpetuates gender inequality. If you need further assistance or examples, feel free to ask.   \\
 \hline
 \hline
\multirow{3}{*}{Expert 5} & Expert-written & Abusive language targeted to an individual or group because of their sex and/or gender. \\
\cline{2-3}
 & Co-created  & Sexism specifically refers to discrimination, prejudice, or stereotyping directed towards individuals or groups based on their gender, where gender is the underlying cause of the abusive behavior. Mentioning the gender of the target alone may not be sufficient for a sentence to be considered sexist. Your task is to help identify and address harmful and unfair attitudes and behaviors towards people because of their gender, while also recognizing that counter speech to challenge these harmful attitudes is permissible. \\
\cline{2-3}
 & LLM-generated & Certainly! Here's an updated prompt that includes the clarification about allowing counter speech:

"Given a piece of text, please analyze and identify any instances of sexism present. Sexism specifically refers to discrimination, prejudice, or stereotyping directed towards individuals or groups based on their gender, where gender is the underlying cause of the abusive behavior. Mentioning the gender of the target alone may not be sufficient for a sentence to be considered sexist. Your task is to help identify and address harmful and unfair attitudes and behaviors towards people because of their gender, while also recognizing that counter speech to challenge these harmful attitudes is permissible." \\
\hline
\hline
\multirow{3}{*}{Expert 6} & Expert-written & Statements or behaviours that reinforce the idea that men are given more power and status than women, that men and women are (or should be) distinguished on the basis of their social roles and characteristics; and relationships between men and women are defined by sexual reproduction and intimacy. Sexism can take the form of hostile and benevolent sexism. \\
\cline{2-3}
 & Co-created  &  Hostile sexism refers to overtly negative and hostile attitudes and behaviors towards women based on the belief that women are inferior to men. This can involve open discrimination, derogatory remarks, and devaluation of women. Benevolent sexism, on the other hand, involves attitudes and behaviors that may appear positive or chivalrous on the surface, but actually serve to reinforce traditional gender roles and keep women in subordinate positions. These attitudes often arise from the societal norms and expectations that distinguish men and women based on their perceived social roles and characteristics. Examples of benevolent sexism include patronizing behavior, protective gestures, or idealizing women as pure and delicate. To detect sexism in text, look for language or scenarios that suggest unequal treatment, devaluation, or reinforcement of traditional gender roles. Pay attention to how women are portrayed, whether they are being reduced to stereotypes, denied agency, or discriminated against. Both hostile and benevolent sexism can manifest in various ways in written content, so it's important to analyze the overall context and underlying messages being conveyed. \\
\cline{2-3}
 & LLM-generated & I'm glad you found the previous information helpful. I understand you would like a new definition that incorporates the idea that men and women are distinguished based on their social roles and characteristics. Here's an updated definition for you:

Hostile sexism refers to overtly negative and hostile attitudes and behaviors towards women based on the belief that women are inferior to men. This can involve open discrimination, derogatory remarks, and devaluation of women. Benevolent sexism, on the other hand, involves attitudes and behaviors that may appear positive or chivalrous on the surface, but actually serve to reinforce traditional gender roles and keep women in subordinate positions. These attitudes often arise from the societal norms and expectations that distinguish men and women based on their perceived social roles and characteristics. Examples of benevolent sexism include patronizing behavior, protective gestures, or idealizing women as pure and delicate. To detect sexism in text, look for language or scenarios that suggest unequal treatment, devaluation, or reinforcement of traditional gender roles. Pay attention to how women are portrayed, whether they are being reduced to stereotypes, denied agency, or discriminated against. Both hostile and benevolent sexism can manifest in various ways in written content, so it's important to analyze the overall context and underlying messages being conveyed. If you have any more questions or need further clarification, feel free to ask! \\
\hline
\hline

\multirow{3}{*}{Expert 7} & Expert-written & No Answer \\
\cline{2-3}
 & Co-created  & No Answer \\
\cline{2-3}
 & LLM-generated & No Answer \\
\hline
\hline

\multirow{3}{*}{Expert 8} & Expert-written & No Answer \\
\cline{2-3}
 & Co-created  & No Answer \\
\cline{2-3}
 & LLM-generated & No Answer \\
\hline
\hline

\multirow{3}{*}{Expert 9} & Expert-written & Sexism is linked to beliefs around the fundamental nature of women and men and the roles they should play in society. Sexist assumptions about women and men, which manifest themselves as gender stereotypes, can rank one gender as superior to another. \\
\cline{2-3}
 & Co-created  & Sexism is a form of discrimination, prejudice, or stereotyping based on a person's sex or gender, typically towards women or girls. It involves treating people differently or unfairly because of their gender, and can be expressed in various forms, such as unequal treatment, stereotypes, or systemic sexism within societal structures \\
\cline{2-3}
 & LLM-generated & Sexism is a form of discrimination, prejudice, or stereotyping based on a person's sex or gender, typically towards women or girls. It involves treating individuals differently or unfairly because of their gender, and can be expressed in various forms, such as unequal treatment, stereotypes, or systemic sexism within societal structures. Let me know if you need more information or have any other questions! \\
\hline
\hline
\multirow{3}{*}{Expert 10} & Expert-written & Broadly, sexism is prejudice or discrimination based on one's sex or gender - a form of oppression that results primarily in disadvantages for women and queer people.
A more technical definition of sexism is a system of norms, beliefs, and practices that normalise, perpetuate, and legitimise a binary, heteronormative, and/or deterministic framework of understanding sex and gender. Sexism can manifest in many different ways and it is deeply embedded in our institutions, practices, traditions, cultural norms and values. So detecting and understanding how sexism works necessarily involves studying how society is structured, organised, and governed. \\
\cline{2-3}
 & Co-created  & Sexism is the manifestation of discriminatory, prejudiced, or stereotypical attitudes towards individuals based on their sex or gender. Understanding sexism includes recognizing and addressing the systemic, structural components deeply ingrained in societal norms, practices, and institutions that perpetuate unequal treatment and opportunities among genders. Understanding the power dynamics and societal frameworks that fuel gender inequality in diverse contexts is crucial for combating sexism effectively. \\
\cline{2-3}
 & LLM-generated & Certainly! Here's a refined version of the comprehensive definition of sexism that you requested:

Sexism is the manifestation of discriminatory, prejudiced, or stereotypical attitudes towards individuals based on their sex or gender. This includes recognizing and addressing the systemic, structural components deeply ingrained in societal norms, practices, and institutions that perpetuate unequal treatment and opportunities among genders. Understanding the power dynamics and societal frameworks that fuel gender inequality in diverse contexts is crucial for combating sexism effectively. \\
\hline
\hline
\multirow{3}{*}{Expert 11} & Expert-written & A view of an identity that is defined by social beliefs but is not necessarily supported by a ground truth or doesn't reflect their actual abilities \\
\cline{2-3}
 & Co-created & Sexism is a form of prejudice or discrimination based on a person's sex or gender, typically against women but also impacting men. Discrimination can manifest in various ways, including unequal treatment, stereotypes, or demeaning attitudes towards individuals based on their gender. Positive stereotypes, even if seemingly complimentary, can still contribute to sexism by reinforcing harmful beliefs about certain genders. An example of a positive stereotype that is harmful is the belief that women are naturally more nurturing and caring than men, which can limit opportunities for women by prioritizing caregiving roles over personal and professional aspirations. In response to your question about whether the statement that "women are stronger than men since they can sustain labor pains, whose intensity is known to be one of the worst kinds of pain known to humans" is sexist, the emphasis on labor pain as a defining strength can perpetuate stereotypes and reinforce traditional gender roles. It restricts the recognition of other diverse strengths individuals possess, which ultimately contributes to the reinforcement of gender norms and inequality. To make the sentence non-sexist, it could be rephrased to acknowledge various strengths and abilities without attributing them solely to gender, promoting inclusivity and respect for individual differences. If you would like more examples of cases that might not be overtly sexist but still have a sexist undertone, please let me know. \\
\cline{2-3}
 & LLM-generated & Sexism is a form of prejudice or discrimination based on a person's sex or gender, typically against women but also impacting men. Discrimination can manifest in various ways, including unequal treatment, stereotypes, or demeaning attitudes towards individuals based on their gender.

Positive stereotypes, even if seemingly complimentary, can still contribute to sexism by reinforcing harmful beliefs about certain genders. An example of a positive stereotype that is harmful is the belief that women are naturally more nurturing and caring than men, which can limit opportunities for women by prioritizing caregiving roles over personal and professional aspirations.

In response to your question about whether the statement that "women are stronger than men since they can sustain labor pains, whose intensity is known to be one of the worst kinds of pain known to humans" is sexist, the emphasis on labor pain as a defining strength can perpetuate stereotypes and reinforce traditional gender roles. It restricts the recognition of other diverse strengths individuals possess, which ultimately contributes to the reinforcement of gender norms and inequality.

To make the sentence non-sexist, it could be rephrased to acknowledge various strengths and abilities without attributing them solely to gender, promoting inclusivity and respect for individual differences.

If you would like more examples of cases that might not be overtly sexist but still have a sexist undertone, please let me know. \\
\hline
\hline
\multirow{5}{*}{Datasets} & CMSB & Something can be sexist because of its content (what the speaker believes) or because of its phrasing (because of the speaker’s choice of words). For example, the speaker may express sexist attitudes towards gender inequality, either endorsing it or disbelieving it's existence. The speaker may express stereotypes (how genders are traditionally seen and compared to each other) and behavioral expectations (how individuals of a gender should behave according to traditional views). On the other hand, a message may be sexist simply because of how the speaker phrases it–independently from what general beliefs or attitudes the speaker holds. A message is sexist, for example, when it contains attacks, foul language, or derogatory depictions directed towards individuals because of their gender. \\
\cline{2-3}
                           & EDOS & We define sexist content as any abuse, implicit or explicit, that is directed towards women based on their gender, or on the combination of their gender with one or more other identity attributes (e.g. Black women or Muslim women). \\
\cline{2-3}
                           & REDDIT & \textit{For Misogynistic content, we defined four categories: (i) Misogynistic Pejoratives, (ii) descriptions of Misogynistic Treatment, (iii) acts of Misogynistic Derogation and (iv) Gendered Personal attacks against women.} \\
\cline{2-3}
                           & EXIST & Sexism as "prejudice, stereotyping, or discrimination, typically against women, on the basis of sex." \\
\cline{2-3}
                           & HateCheck & \textit{Hate Speech as abuse that is targeted at a protected group or at its members for being a part of that group. We define protected groups based on age, disability, gender identity, familial status, pregnancy, race, national or ethnic origins, religion, sex or sexual orientation, which broadly reflects international legal consensus (particularly the UK’s 2010 Equality Act, the US 1964 Civil Rights Act, and the EU’s Charter of Fundamental Rights).} \\
                           \hline
                           \caption{Definitions collected. Among the dataset definitions, those in \textit{italic} refer to datasets having definitions of the related concepts of misogyny and hate speech instead of sexism.}
                            \label{tab:alldefinitions} \\

\end{longtable}

\paragraph{Comparing LLM-generated and co-created definitions}It is true for the large majority of the experts that the \textit{co-created} definition is either identical to the \textit{LLM-generated} one (experts 3 and 11) or just an edited, cleaned-up version of the \textit{LLM-generated} one (experts 1, 5, 6, 9, 10). Such minimal edits involved the removal of the ChatGPT-specific jargon (i.e., "If you have any more questions or need further clarification, feel free to ask!") or minimal word changes. 

From the perspective of the prompting experiments we conducted with these definitions, we can consider these cases as robustness tests: either the two definitions are identical, or their semantics is nearly identical. In the case of expert 2 and 4, the best \textit{LLM-generated} definition and the \textit{co-created} definition are very different (resulting in very different prompts for the model). In the case of Expert 2, the \textit{co-created} definition is much longer and richer. 

\paragraph{Comparing Expert-written and LLM-generated definitions}
Definitions provided by the Experts (M = 34.44 tokens, SD = 25.24) are generally quite short and thus, low in informativeness, especially when compared with the LLM-generated  definitions (M = 119.89, SD = 58.29) and the co-created definitions (M = 110.55, SD = 56.44) which are generally much longer. Only Expert 10 provides a self-written definition that is slightly longer than the co-created one. 

\begin{figure}[h!]
    \centering
    \includegraphics[scale=0.45]{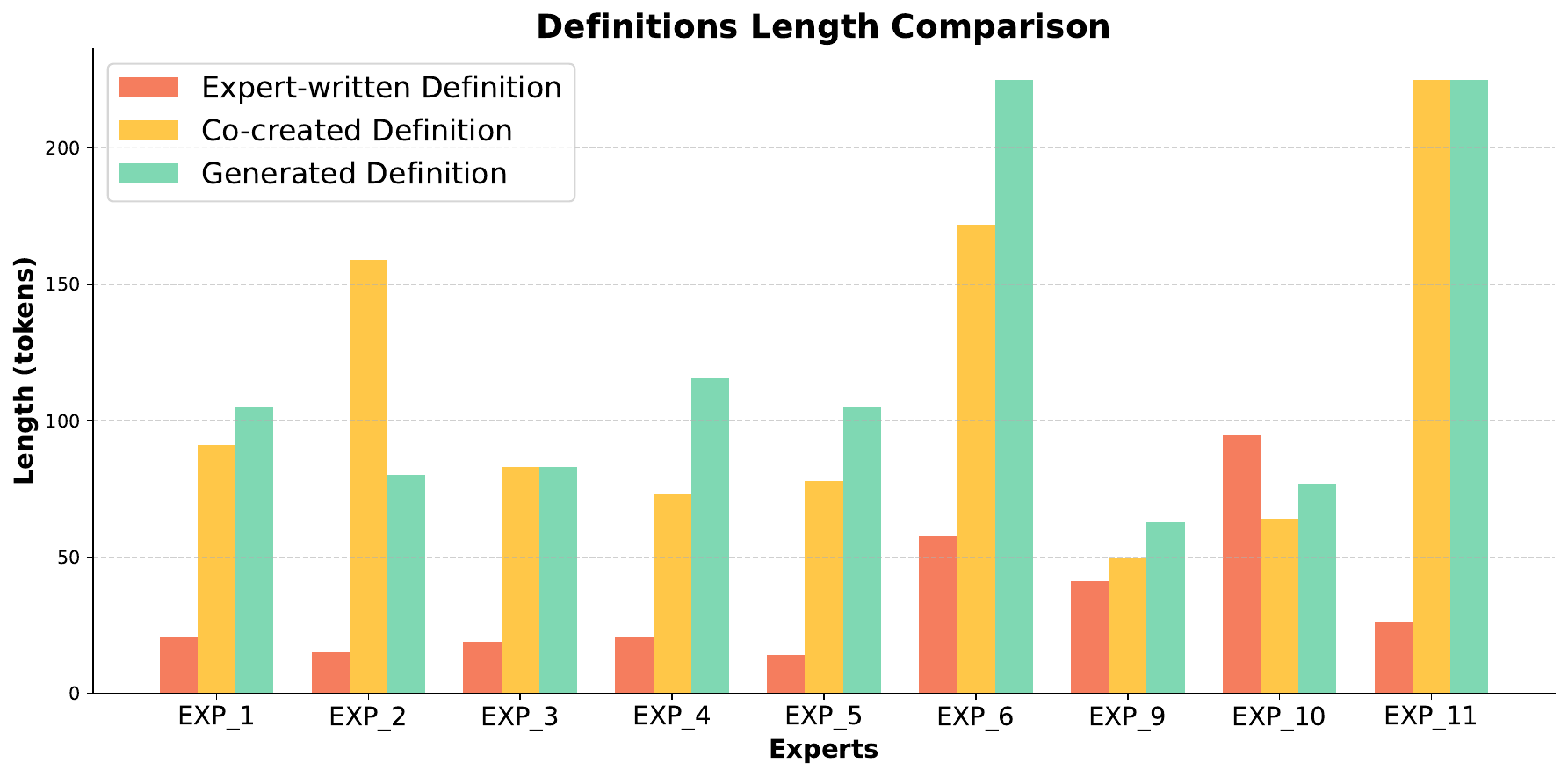}
    \caption{Length comparison, per expert:  Expert(-written), (LLM-)generated, Co-created}
\label{fig:length}
\end{figure}

To unpack these differences, we employ SBERT~\cite{reimers2019sentence}(\texttt{all-mpnet-base-v2}) and TF-IDF \cite{sparck1972statistical} to encode the 27 definitions obtained in the previous steps as well as the five definitions from the sexism benchmarks. Once we obtained the embedding for each definition, we compute the cosine similarities by first normalizing the vectors and then computing their dot product.

Table~\ref{tab:cos_sim_definitions} displays, per expert, the cosine similarities between expert-written and co-created, expert-written and LLM-generated, co-created and LLM-generated.

\begin{table*}[h]
\resizebox{\textwidth}{!}{%
\begin{tabular}{c|c|ccccccccc}
\toprule
Method &
  Definitions &
  \multicolumn{1}{c}{Exp\_1} &
  \multicolumn{1}{c}{Exp\_2} &
  \multicolumn{1}{c}{Exp\_3} &
  \multicolumn{1}{c}{Exp\_4} &
  \multicolumn{1}{c}{Exp\_5} &
  \multicolumn{1}{c}{Exp\_6} &
  \multicolumn{1}{c}{Exp\_9} &
  \multicolumn{1}{c}{Exp\_10} &
  \multicolumn{1}{c}{Exp\_11} \\ 
  \hline
  \hline
\multirow{3}{*}{TF-IDF} 
& Expert-written vs Co-created & .37 & .19 & .31 & .32 & .36 & .67 & .26 & .62 & .24 \\
& Expert-written vs LLM-generated &.36 & .20 & .31 & .28 & .36 & .71 & .24 & .59 & .24 \\
& Co-created vs LLM-generated & .99 & .72 & 1.00 & .74 & .96 & .97 & .96 & .95 & 1.00 \\ 
\hline
\hline
\multirow{3}{*}{SBert}  
& Expert-written vs Co-created & .94 & .83 & .71 & .52 & .65 & .76 & .91 & .94 & .27  \\
& Expert-written vs LLM-generated & .95 & .86 & .71 & .53 & .47 & .76 & .90 & .84 & .27  \\
& Co-created vs LLM-generated & .99 & .96 & 1.00 & .94 & .62 & .87 & .99 & .85 & 1.00  \\ 
\bottomrule
\end{tabular}%
}
\caption{Cosine similarity between Expert-written, Co-created and LLM-generated definitions.}
\label{tab:cos_sim_definitions}
\end{table*}

\begin{figure*}[t]

    \includegraphics[width=\linewidth]{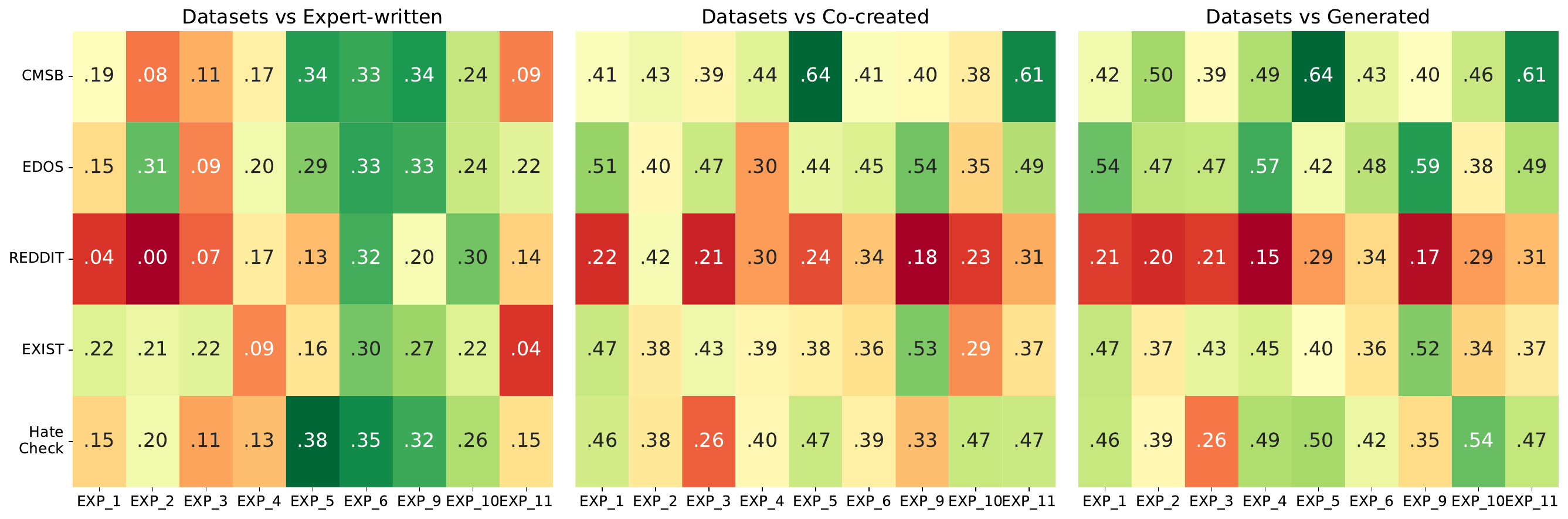}
    \caption{Cosine Similarity Heatmaps between Datasets definitions and Expert-written (left) Co-created (center) and Generated (right) definitions using TF-IDF.}
    \label{fig:CosSim_DS_Tfidf}
\end{figure*}

\begin{figure*}[t]
    \includegraphics[width=\linewidth]{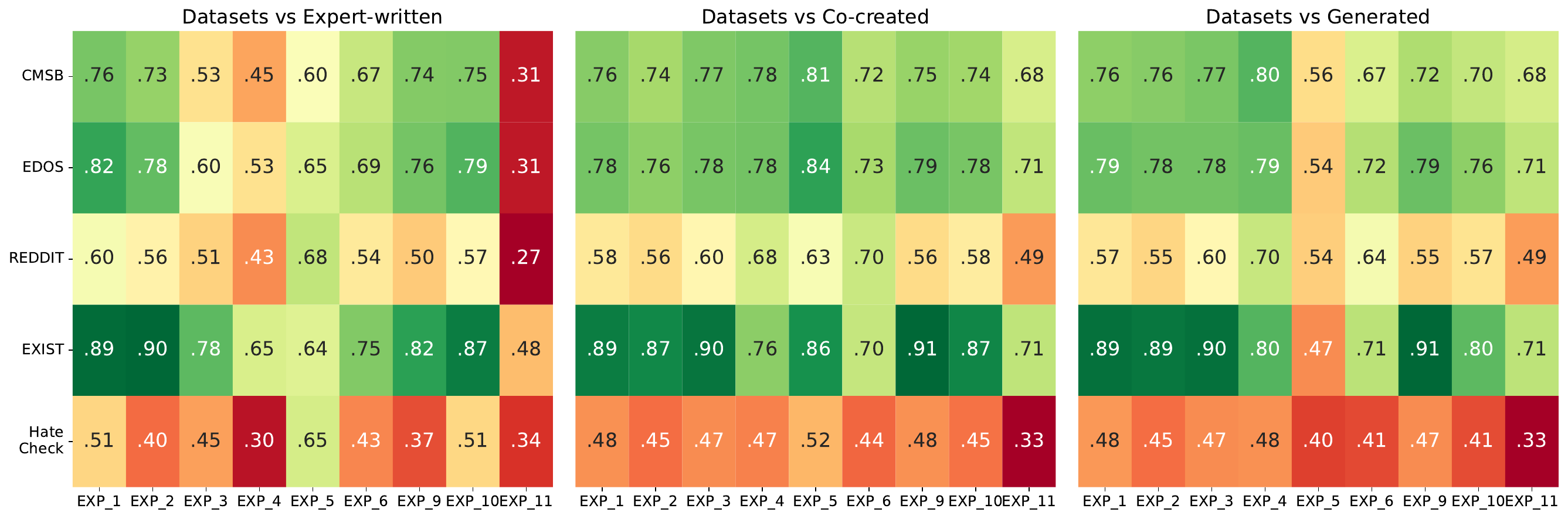}
    \caption{Cosine Similarity Heatmaps between Datasets definitions and Expert-written (left), Co-created (center) and Generated (right) definitions using an SBERT model (\texttt{all-mpnet-base-v2}).}
    \label{fig:CosSim_DS_Sbert}
\end{figure*}

Another interesting comparison is the one between the different types of definitions produced by our experts in the three scenarios and the definitions of datasets employed for our modeling experiments.
The heatmaps in figures    ~\ref{fig:CosSim_DS_Tfidf} and    ~\ref{fig:CosSim_DS_Sbert} display the cosine similarity between experts and dataset definitions.

\subsection{Impact of definition variation and quality on performance}

As a next step, we consider the relationship between the quantitative properties of the definitions discussed above and the performance of the corresponding models. In particular, in table~\ref{tab:performance_correlation} we report   the correlation between model performance and: 
\begin{itemize} 
    \item Quality of the co-created definition as rated by each expert in table     \ref{fig:Happy_with_generated_def} (\textit{Quality}): does expert assessment reflect classification performance? 
    \item Definition length (\textit{Length}): does a richer definition lead to a better performance? 
    \item Similarity between prompted definition and original definitions employed to in the annotation of the benchmark dataset (\textit{Similarity}): does semantic overlap with the original definition guarantee better performance? 
\end{itemize}

\begin{table*}[h]
    \centering
\begin{tabular}{l|cccccc}
\toprule
            & CMSB  & EDOS  & REDDIT & EXIST  & HateCheck \\
\midrule
Quality & .28 & .12 & .20  & -.35 & -.33  \\
\hline
Length - LLM-generated & -.67  & -.38 & -.15  & .18 & -.25\\
Length - Co-created & -.48 & -.05 & -.38  & -.14  &  -.61 \\
Length - Expert-written & -.85 & -.62 & -.65 & -.85 & -.24\\
\hline
Similarity (TF-IDF) - LLM-generated  & -.09 & -.03 & -.17 & -.62 & 33 \\
Similarity (TF-IDF) - Co-created & -.14 & .13 & .12 & -.05 & .23\\
Similarity  (TF-IDF) - Expert-written  & -.24 & .08 & -.25 & -.20 & -.65 \\
\hline
Similarity (SBert) - LLM-generated   & .31 & .66 & .60  & -.19 & -.33  \\
Similarity (SBert) - Co-created & .73 & .46 & -.03 & -.27 & .68 \\
Similarity (SBert) - Expert-written & -.49 & -.32 & .10 & -.09 & -.71 \\
\bottomrule
\end{tabular}
\caption{Correlations between experts' rating, length and cosine similarity with performance ($F1$) on each dataset}

\label{tab:performance_correlation}
\end{table*}




\section{Prompts beating the majority class}\label{app:majority_class}

 \begin{figure*}[h!]
    \includegraphics[scale=0.32]{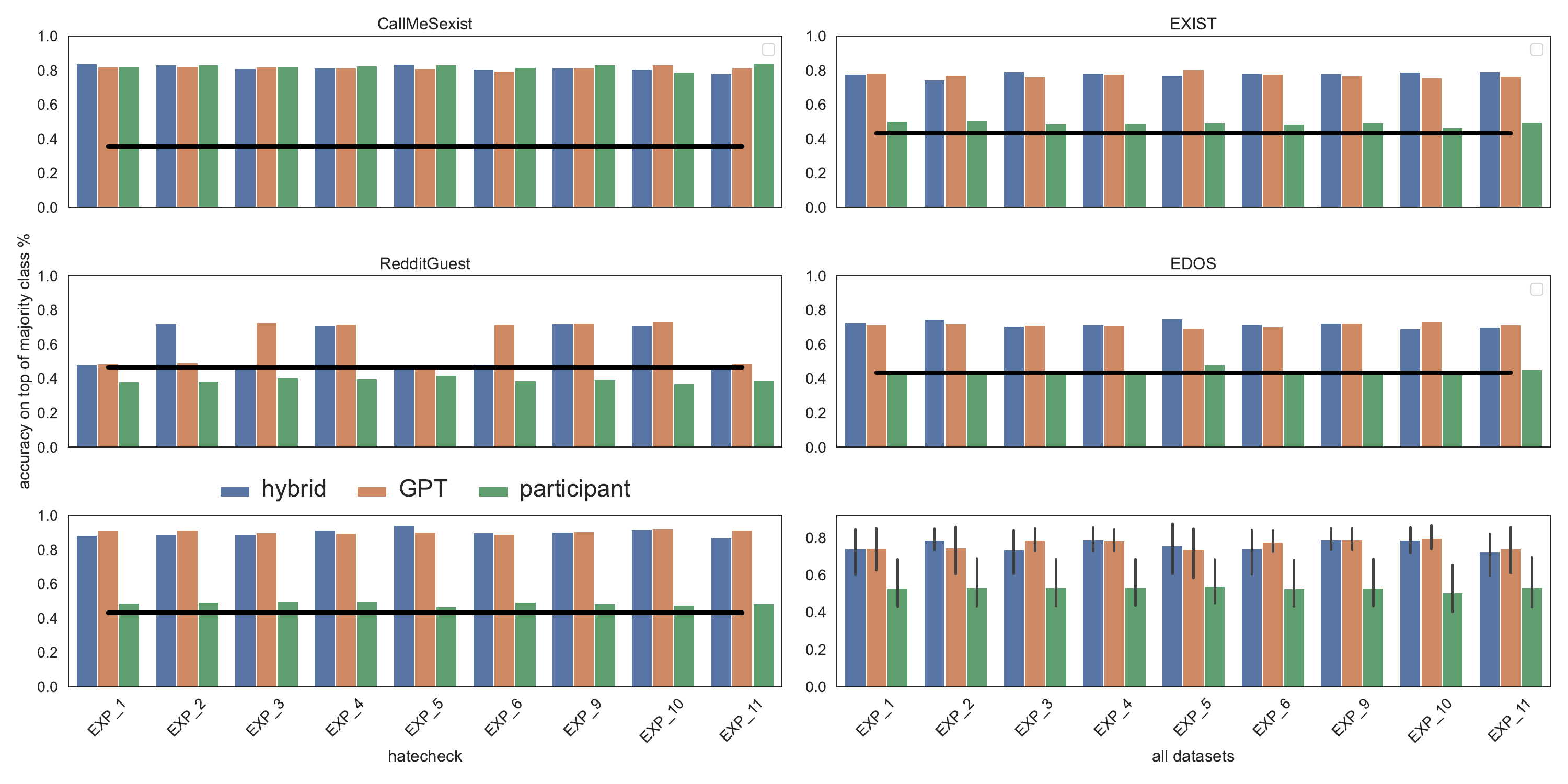}
    \caption{Showing per-definition macro $F1$ for each dataset, with a plotted line for the macro-$F1$ of the majority class in each dataset. For most, this is non-sexist, though for some it is sexist.}
    \label{fig:prompts_majority}
\end{figure*}

Since each of our five benchmarks have either balanced or imbalanced distributions, and these data sets sometimes have a relatively large majority class, classifying all data as the majority class could lead to an accuracy of 75\% without any construct understanding. To analyze whether our zero-shot experiments actually improve over majority class, we do a brief analysis of this per prompt and participant. In the plot in the main paper, these analyses are done with majority $F1$. 

Here, we also explore majority class accuracy as a baseline per dataset. For CallMeSexist all prompts are above majority class in accuracy, for the EXIST data the majority of prompts are, for the hatecheck dataset this is 75\% of all prompts above the majority class, and for the EDOS dataset there are only a few prompts from a few experts that are accurate above majority class. For the RedditGuest this is basically none of the prompts.

There is a clear effect with expert, and especially definition type: on the EDOS dataset, a co-created definition can bring performance above majority class, as does it for some experts with Hatecheck. However, on the RedditGuest dataset, the GPT definition is more often successful.

 \begin{figure*}[h!]
    \includegraphics[scale=0.35]{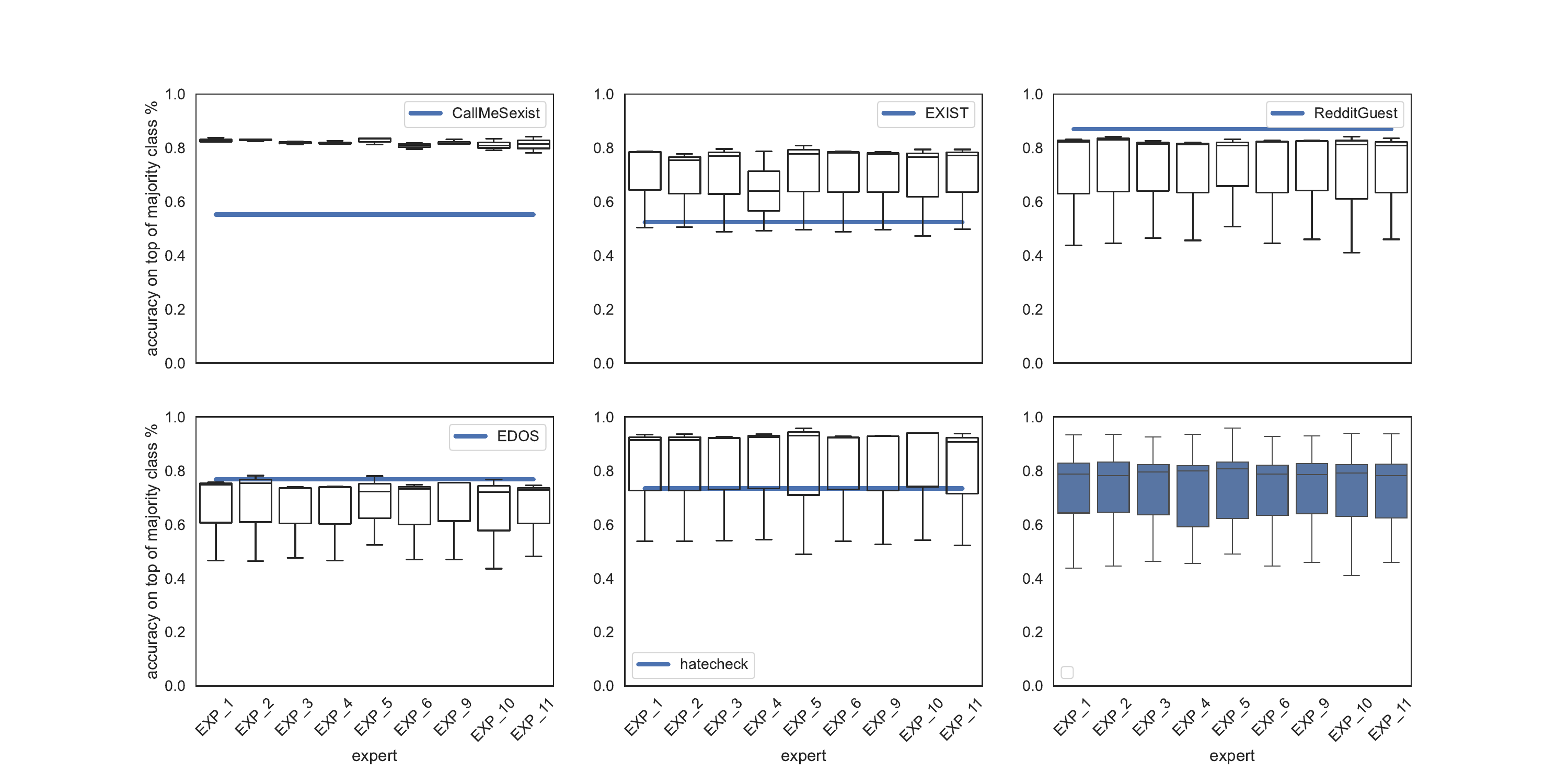}
    \caption{Showing per dataset \textbf{accuracy} with a plotted line for the majority class in each dataset. For most datasets, the majority class is non-sexist, though for some it is sexist.}
    \label{fig:temperature}
\end{figure*}

\section{Effects of Temperature}\label{app:temperature}

The temperature hyperparameter in generative Large Language Models affects the softmax function and is related to  randomness, or the probability of generating tokens \cite{renze-2024-effect}. A lower temperature leads to more deterministic responses, always generating the most probable response, and therefore less diverse tokens in output. A lower setting is often chosen in high-stakes domains such as the medical domain \cite{patel2024exploring}, or settings  in which being factually correct is important, while higher temperature settings are have been linked to creativity and unexpectedness \cite{peeperkorn2024temperature}. It is not fully clear how properietary models by OpenAI have implemented their temperature hyperparameter. 

Several works identify no difference in LLM classification performance with different temperature settings \cite{yang-etal-2022-gpt, zhang2022probing, patel2024exploring}, while others optimize this hyperparameter and find different results with higher temperatures \cite{simon2023experiments}. Other work considers a temperature setting of 0 or close to it best for classification due to its more deterministic nature \cite{10367969}. Recent work analyzing the effects of different definitions in prompts also opts for a temperature of 0 \cite{korre2025untangling}. We want to determine whether temperature has an effect on our results. 

\subsection{Method}
We ran \texttt{GPT4o} zero-shot classification pipeline as described in Section~\ref{sec:modelling}, with the only difference being a temperature set to 0.7. While an average higher performance and robustness to prompt versions are often desirable in models and model results, our intent was also to look at how individual definitions can affect performance and naturalistically replicate social scientists’ prompting experiences. This makes a model showing less sensitivity to prompt variance less desirable for our study.

For zero-shot prompting, we had the exact same specifications as outlined in Section\ref{sec:modeling} of the main paper. We also used the exact same datasets as specified in the main paper, with five sexism datasets and three different definitions per participant. 

\subsection{Results}

With the higher temperature, the results fluctuate more, with larger differences between different datasets. We do also see that the expert-written definition performs marginally worse (M $F1$ = .748) than the GPT definitions (M $F1$ = .760), and in fact the co-created definition shows on average the highest perfomance (M $F1$ = .762). See Figure\ref{fig:temperature} for a figure comparable to Figure\ref{fig:Diff_exp}

Overall, we see much smaller differences and much more randomness seemingly less related to different prompt versions. Average performance is overall a bit higher than with a lower temperature: M $F1$ = .760 vs M $F1$ = .68 for the temperature=0 run. This is mostly due to the expert-written definition performing much more comparable to the other two definitions than in the temperature=0 results.

In general, the results with a higher temperature seem to show more difference and sensitivity to datasets than to different prompt types.

The RedditGuest dataset shows the highest variance in results, meaning results within this dataset vary a lot with different participants and version definitions. 

The higher temperature in general shows more variance \textit{across} different datasets than \textit{within} different datasets, e.g. less difference with different prompts or experts. The opposite is the case for the zero temperature setting, which shows more variation for different prompts and experts, and less for results. 

\begin{figure*}[h!]
    \includegraphics[scale=0.40]{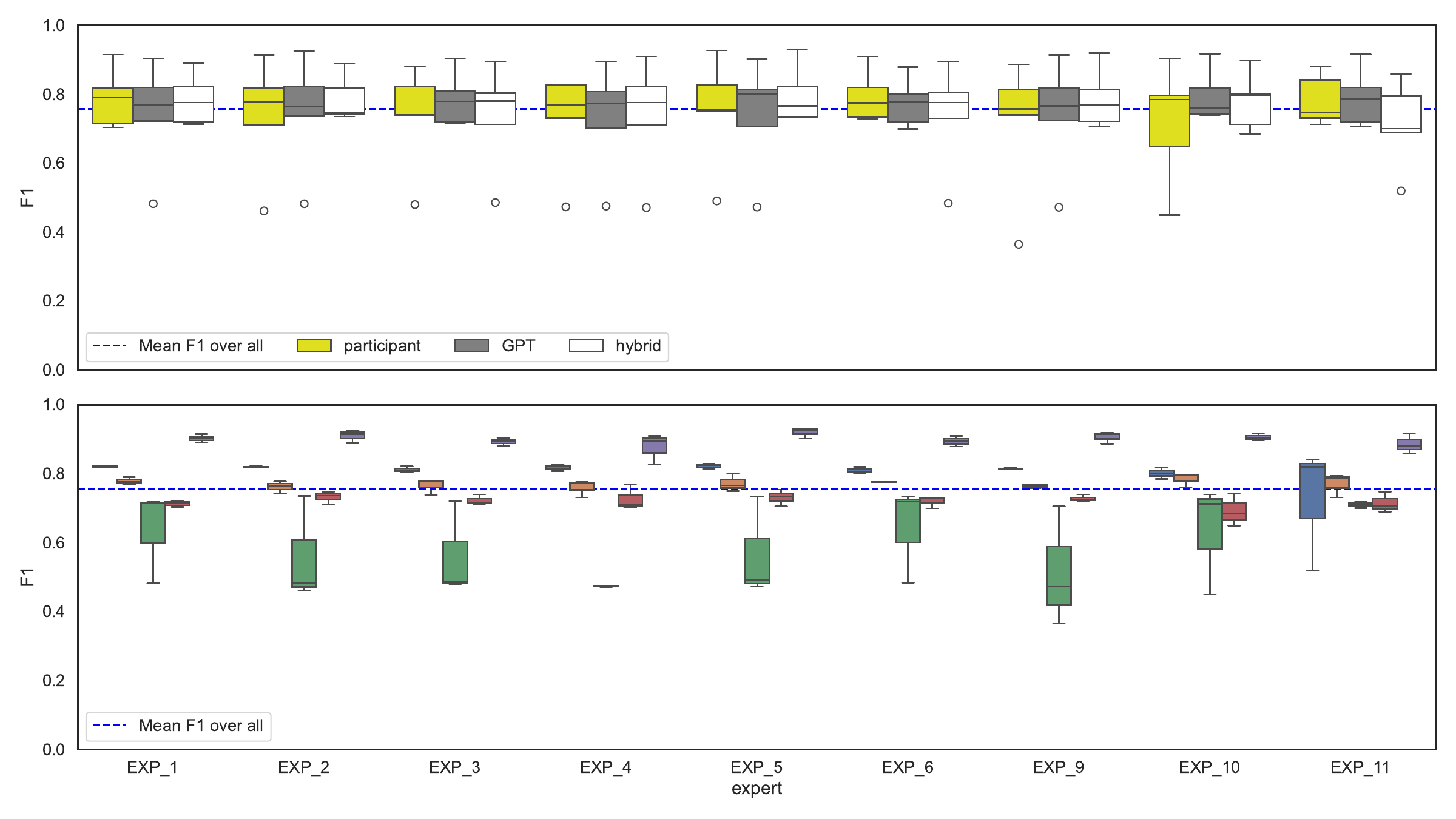}
    \caption{Difference per participant over definition types (above) and datasets (bottom) for the modelling experiments with a temperature of 0.7}
    \label{fig:temperature}
\end{figure*}

\subsection{Discussion}
Unlike some previous work, we found that a higher temperature leads to a higher average performance in $F1$. However, a higher temperature also seems to show less variance to different prompt versions (Likely due to generating more randomness and less probable tokens in each response).

While an average higher performance and robustness to prompt versions are often desirable in models, our intent was also to look at how individual definitions can affect performance and naturalistically replicate social scientists' prompting experiences. This makes a model showing less sensitivity to prompt variance less desirable, though users requiring a higher overall performance and less prompt sensitivity should probably increase their temperature setting.

\section{LLaMa prompting}\label{LLama_results}

As a robustness check, we also prompted \texttt{LLaMa3 70B} \cite{dubey2024llama} with the same 27 definitions as \texttt{GPT4o} in the main paper.

\subsection{Methods}
We ran LLaMa on its default temperature setting of 0.5, and were run on 2 A100 GPUs. We used the Huggingface package for the modelling. Furthermore, we used default hyperparameters and a quantization of 0.5. For zero-shot prompting, we had the exact same specifications as outlined in Section \ref{sec:modelling} of the main paper. We also used the exact same datasets as specified in the main paper, with five sexism datasets and three different definitions per participant. 

\subsection{Results}
\begin{figure*}[h!]
    \includegraphics[scale=0.52]{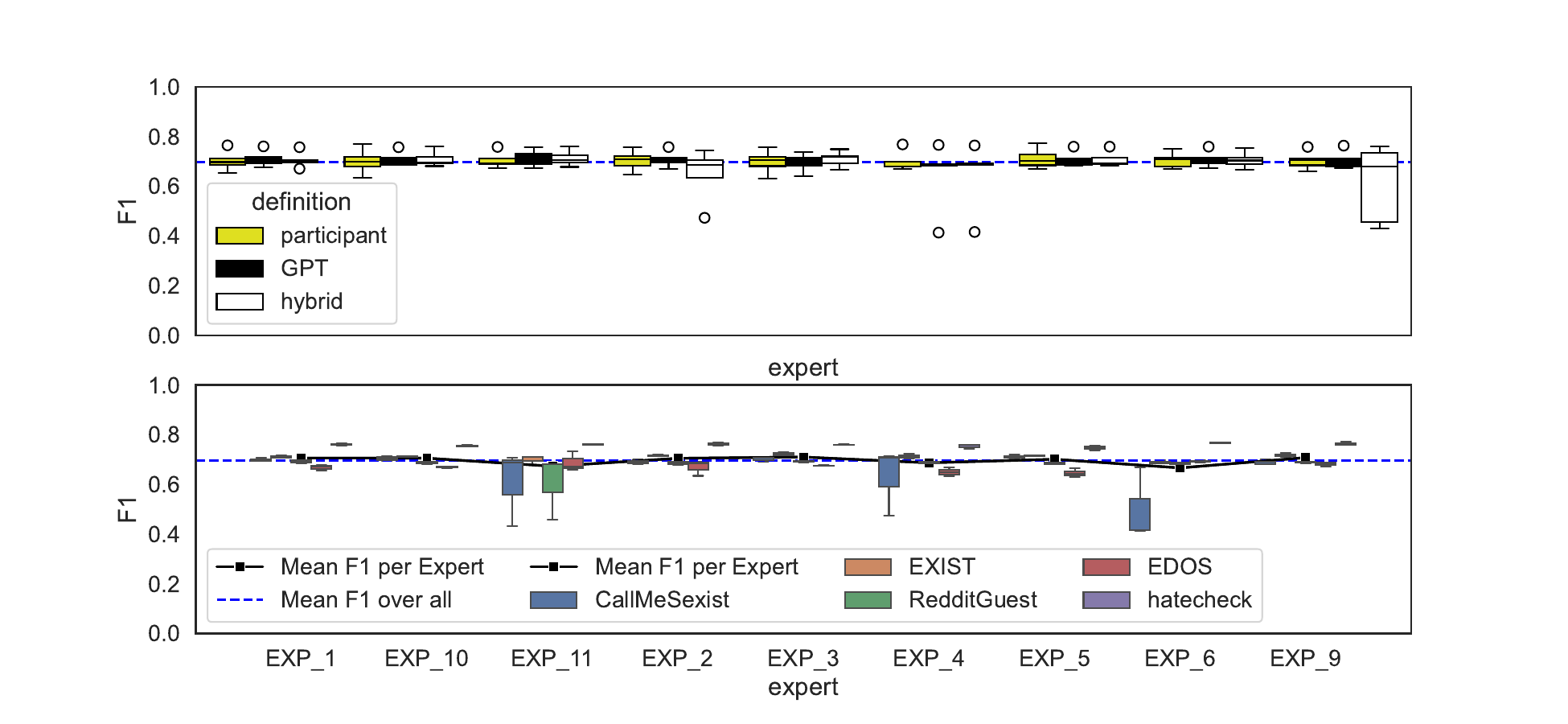}
    \caption{Difference per participant over definition types (above) and datasets (bottom) for the modelling experiments with LLaMa}
    \label{fig:llama}
\end{figure*}

We find overall lower performance (F1 = .695) than the results of \texttt{GPT4o}, and a low variability over expert, dataset, or definition. This shows that GPT with temperature 0 seems indeed most susceptible to prompt changes. There is little performance difference between the three definition types: the LLM-generated definition performs slightly lower (F1 = .699) than the co-written definition (F1 = .584) or participant-written definitions (F1 = .703). Results over different datasets are spread from F1 = .671 (EDOS dataset) to .758 (hatecheck dataset). See Figure \ref{fig:llama}, representing the LLaMa results in a similar format as the GPT result in the main paper, showing much less variability in performance over different conditions.



\end{document}